%% file: icml_poisoning.tex
\documentclass[nohyperref]{article}

\usepackage{microtype}
\usepackage{graphicx}
\usepackage{subcaption}
\usepackage{booktabs} 

\usepackage{hyperref}

\usepackage[accepted]{icml2022}

\usepackage{amsmath}
\usepackage{amssymb}
\usepackage{mathtools}
\usepackage{amsthm}
\usepackage{bbm}
\usepackage{ulem}

\usepackage[capitalize,noabbrev]{cleveref}

\theoremstyle{plain}
\newtheorem{theorem}{Theorem}[section]

\theoremstyle{definition}

\theoremstyle{remark}

\usepackage{xspace}
\newcommand{\alg}{\textsc{EPIc}\xspace}

\usepackage[textsize=tiny]{todonotes}

\icmltitlerunning{Not All Poisons are Created Equal: Robust Training against Data Poisoning}

\begin{document}

\twocolumn[
\icmltitle{Not All Poisons are Created Equal: Robust Training against Data Poisoning}



\icmlsetsymbol{equal}{*}

\begin{icmlauthorlist}
\icmlauthor{Yu Yang}{ucla}
\icmlauthor{Tian Yu Liu}{ucla}
\icmlauthor{Baharan Mirzasoleiman}{ucla}
\end{icmlauthorlist}

\icmlaffiliation{ucla}{Department of Computer Science, University of California, Los Angeles, United States}

\icmlcorrespondingauthor{Yu Yang}{yuyang@cs.ucla.edu}

\icmlkeywords{Machine Learning, ICML}

\vskip 0.3in
]



\printAffiliationsAndNotice{}  

\begin{abstract}
\input{abs}

\end{abstract}
\input{intro}

\input{related}
\input{method}
\input{analysis}
\input{experiments}
\vspace{-2mm}
\input{conclusion}

\vspace{-3mm}
\section*{Acknowledgements}
\vspace{-2mm}
This research was supported in part
by Cisco Systems and UCLA-Amazon Science Hub for Humanity and AI.


\bibliography{icml_poisoning}
\bibliographystyle{icml2022}

\newpage
\appendix
\onecolumn
\input{appendix}

\end{document}

%% file: abs.tex
Data poisoning 
causes misclassification of test time target examples, 
by injecting maliciously crafted samples in the training data.
%
Existing defenses are often effective only against a specific type of
{targeted} attack, significantly degrade the generalization performance, or are prohibitive for 
standard deep learning pipelines.
In this work, we propose an efficient defense mechanism that significantly reduces the success rate of various data poisoning attacks, and provides theoretical guarantees for the performance of the model.
Targeted attacks work by adding bounded perturbations to a randomly selected
subset of training data to match the 
targets' gradient or representation.
We 
show that:
(i) 
under bounded perturbations, only a number of poisons can 
be optimized to have a gradient that is close enough to that of the target and make the attack successful; 
(ii) such effective poisons move away from their original class and get isolated 
in the gradient space;
(iii) dropping examples 
in low-density gradient regions
during training can successfully eliminate the effective poisons, 
and guarantees similar training dynamics to that of training on full data.
Our extensive experiments show that our method significantly decreases the success rate of state-of-the-art targeted attacks, including Gradient Matching and Bullseye Polytope, and easily scales to large datasets\footnote{Code is available at \href{https://github.com/YuYang0901/effective-poison-identification}{https://github.com/YuYang0901/Epic}.}.


%% file: intro.tex
\vspace{-5mm}\section{Introduction}
The impressive success of modern machine learning systems is highly dependent on the quality of their large training data. Many large datasets are scraped from the internet, or other public and user-provided sources. Models trained on such datasets are susceptible to data poisoning attacks, wherein an adversary places specially-constructed poisoned examples into the training data with the intention of manipulating the behavior of the system at test time. 
These attacks create security vulnerabilities that cannot be detected even if the data is labeled and checked by human supervision.
This makes data poisoning arguably one of the most concerning threats to deep learning systems
deployed in security- and safety-critical applications, such as financial services, security cameras,  autonomous cars, and medical devices. 

Various types of poisoning attacks have been proposed in recent years. {Most attacks fall into one of two main categories: backdoor or triggerless poisoning.} Backdoor data poisoning augments the training data by a set of poisoned examples that contain a (not necessarily visible) trigger pattern \cite{gu2017badnets,turner2018clean,souri2021sleeper}. 
Finetuning the model on the augmented training data causes a model to misclassify test-time samples containing the trigger. 
On the other hand, triggerless poisoning attacks work by crafting {small per-example perturbations so that the perturbed training examples} collide with the {adversarially labeled} target in the feature or gradient space
\cite{Shafahi2018poisonfrogs,zhu2019transferable,huang2020metapoison,geiping2021witches,aghakhani2021bullseye}.
Triggerless poisoning attacks cause misclassification of particular instances and do not require modification at inference time. 
In both cases, the poisoned examples may be seemingly innocent and properly labeled, and hence are hard to be detected by expert observers. 

Existing defense mechanisms {against data poisoning attacks} mainly rely on either anomaly detection based on nearest neighbors, training loss, singular-value decomposition, feature and activation clustering \cite{cretu2008casting, Steinhardt17certified,tran2018spectral,chen2019detecting,peri2020deep}, or robust training based on strong data augmentation, randomized smoothing, ensembling, and adversarial training \cite{weber2020rab,levine2020deep,abadi2016deep,ma2019data,li2021anti,tao2021better}.   
However, such methods either drastically degrade the generalization performance of the model \cite{geiping2021doesn}, or can only protect the model against certain types of poisoning attacks \cite{koh2018stronger,tran2018spectral}, or 
are {computationally} prohibitive for standard deep learning pipelines \cite{geiping2021doesn}. Importantly, these methods do not provide any theoretical guarantee for the performance of the model \cite{weber2020rab,levine2020deep,abadi2016deep,geiping2021doesn}.

We develop an efficient and principled defense framework that effectively prevents various types of targeted poisoning attacks, 
and provide theoretical guarantee for the performance of the model. 
To successfully prevent poisoning attacks, we make the following key observation: not all poisons are effective in making the attack successful. In particular, targeted attacks add bounded perturbations to randomly selected subsets of training data to match the gradient of the adversarially labeled target.
We show that for a poison to be {effective}, it needs to fall close enough to the target in the gradient space. 
However, under bounded perturbations, only a small number of poisons can be optimized to get close enough to the target and make the attack successful. Such effective poisons get far away from their original class and get isolated in the gradient space. Eliminating the \textit{effective poisons} can successfully break various types of attacks.\looseness=-1

To prevent data poisoning while maintaining the generalization performance of the network, we aim to identify and eliminate the effective poisons.
We show that effective poisons can be identified as isolated
\textit{medoids} of each class, in the gradient space.
Medoids are the most centrally located examples of a dataset, that minimize the sum of dissimilarity between every data point to its nearest medoid.
The set of medoids can be efficiently extracted by maximizing a submodular function.
To eliminate effective poisons, we iteratively find 
medoids of every class in the gradient space during the training. 
Then, we assign every data point to {the} closest medoid {in its class}, and drop the medoids to which no other data point is assigned.
We show that our Effective Poison IdentifiCation (\alg) method can successfully eliminate effective poisons. 
We also prove that training on large gradient clusters of each class guarantees similar training dynamics to that of training on the full data. 

Compared to existing defense strategies, our method does not require a pre-trained clean model, is not attack-specific, can be applied very efficiently during the training, and provides a quality guarantee for the performance of the trained model.
Our extensive experiments show that our method
renders state-of-the-art
targeted attacks, including Gradient Matching,
Bullseye Polytope, and Feature Collision ineffective, with only a slight decrease in the performance. 
We note that, \alg~is the only effective defense method against state-of-the-art attacks that can efficiently scale to standard deep learning pipelines. Compared to the state-of-the-art \citep{geiping2021doesn}, \alg~is 6.9x faster, and maintains similarly high test accuracy and low attack success rate.


%% file: related.tex
\vspace{-3mm}
\section{Related Work}
\subsection{Targeted Data Poisoning}

Attacks on deep networks can be generally divided into triggered and triggerless attacks. Triggered or backdoor attacks augment the training data with a small set of examples that contain a trigger {patch} and belong to a specific target label. 
Models trained on the augmented data will misclassify test examples with the same patch. While early backdoor attacks were not clean-label \cite{chen2017targeted, gu2017badnets, liu2017trojaning,souri2021sleeper}, recent backdoor attacks produce poison examples that do not contain a visible trigger \cite{turner2018clean, Saha2019htbd}.
triggerless poisoning attacks add small adversarial perturbations to base images to make their feature representations or gradients match that of the {adversarially labeled} target \cite{Shafahi2018poisonfrogs,zhu2019transferable,huang2020metapoison,geiping2021witches,aghakhani2021bullseye}. Such poisons are very similar to the base images in the input space, cannot be detected by observers, and
do not require modification to targets at inference time. 
The most prominent poisoning attacks 
we test our defense against are:

\vspace{-1mm}
\textbf{Feature Collision (FC)} crafts poisons by adding small perturbations to base examples so that their feature representations collide with that of the target \cite{Shafahi2018poisonfrogs}. 

\vspace{-1mm}
\textbf{Bullseye Polytope (BP)} is similar to FC, 
but instead crafts poisons such that the target resides close to the center of their convex hull in feature space \cite{aghakhani2021bullseye}.



\vspace{-1mm}
\textbf{Gradient Matching (GM)} 
produces poisons by approximating this bi-level objective using ``gradient alignment", encouraging gradients of the clean-label poisoned data to align with that of the adversarially labeled target \cite{geiping2021witches}. This attack is shown to be effective against data augmentation and differential privacy.



\vspace{-1mm}
{\textbf{Sleeper Agent (SA)}} is a hidden-trigger backdoor attack that also craft poisons based on the ``gradient alignment"  between patched poisons and targets \cite{souri2021sleeper}.\looseness=-1

\vspace{-2mm}
\subsection{Defense Strategies}
\vspace{-2mm}
Commonly used data sanitization defenses 
work by detecting anomalies 
that fall outside a spherical radius
in the feature space \cite{Steinhardt17certified}, spectrum of the feature covariance matrix \cite{tran2018spectral}, or activation space \cite{chen2019detecting}. They may also filter points that are labeled differently from their nearest neighbors in the feature space \cite{peri2020deep}. 
Such defense mechanisms rely on the assumption that poisons are far from the clean data points in the input or feature space. Hence, they can be easily broken by stronger data poisoning attacks that place poisoned points near one another, or by optimization methods that craft poisons to evade detection \cite{koh2018stronger,Shafahi2018poisonfrogs,Saha2019htbd}.

Robust training methods rely on strong data augmentation \cite{borgnia2021strong}, apply randomized smoothing \cite{weber2020rab}, use an ensemble of models for prediction \cite{levine2020deep}, or
bound gradient magnitudes and minimize differences in orientation \cite{hong2020effectiveness}.
Such methods often incur a significant performance penalty \cite{jayaraman2019evaluating}, and  can even be {adaptively attacked} by modifying gradient signals during poison crafting \cite{veldanda2020evaluating}.
Other 
identify backdoor attacks early in training and revert their effect by gradient ascent \cite{li2021anti}, use adversarial training \cite{madry2018towards,tao2021better}, 
or create poisons during training and inject them into
training batches
\cite{geiping2021doesn}.

Existing defense methods either drastically degrade the model's performance \cite{geiping2021doesn}, only protect the model against certain types of poisoning attack \cite{koh2018stronger,tran2018spectral}, are
prohibitive for larger datasets \cite{geiping2021doesn}, or do not provide any theoretical guarantee for the performance of the model \cite{weber2020rab,levine2020deep,abadi2016deep,geiping2021doesn}.
On the other hand, our method is fast and scalable, and 
successfully eliminates various poisoning attacks while allowing the model to learn effectively from clean examples with rigorous theoretical guarantees.

%% file: method.tex
\vspace{-2mm}
\section{Robust Training against Data Poisoning}


Let $\mathcal{D}_c=\{(x_i,y_i)\}_{i=1}^{n}$ be the set of all clean training data, where $x_i\in\mathbb{R}^m$. 
Targeted data poisoning attacks 
aim to change the prediction of a target image $x_t$ in the test set to an adversarial label $y_{\text{adv}}$, 
by modifying a fraction (usually less than 1\%) of data points 
in the training data  within an $l_\infty$-norm $\epsilon$-bound.
We denote by $V=\{1,\cdots,n\}$, and $V_p\subset V$ the index set of the entire training data and poisoned data points, respectively.
For small $\epsilon$,  this constraint enforces
the perturbed images to look visually similar to the original example. Such attacks remain visually invisible to human observers and are called clean-label attacks.
%
Targeted clean-label data poisoning attacks can be formulated as the following bi-level optimization problem:
\begin{align}\label{eq:poisoning}
    \min_{\delta \in\mathcal{C}} \mathcal{L}(x_t, y_{\text{adv}}, \theta(\delta)) &\quad s.t.\quad\\ &\theta(\delta)\! =\! {\arg\min}_\theta \sum_{i\in V} \mathcal{L}(x_i\!+\!\delta_i, y_i, \theta),\nonumber
\end{align}
where $\mathcal{C}\!\!=\!\{\delta\!\in\mathbb{R}^{n\times m}\!\!: \!\|\delta\|_\infty\!\!\leq\epsilon, \delta_i\!=\!0 ~\forall i\notin V_p\}$ is the constraint set determining the set of valid poisons. 
Intuitively, the perturbations change the parameters $\theta$ of the network such that  minimizing the training loss on RHS of Eq. \!\!\eqref{eq:poisoning}
also minimizes the adversarial loss on LHS of Eq. \!\! \eqref{eq:poisoning}.

{{
We assume
that the network is trained by minimizing the training loss $\mathcal{L}(\theta)=
\sum_{i\in V}\mathcal{L}(x_i+\delta_i,y_i, \theta)$
over the entire set of clean and poisoned training examples $i\!\in\! V, \delta_i\!=\!0 ~\forall i\notin V_p$. 
Applying gradient descent with learning rate $\eta$ to minimize the training loss $\mathcal{L}(\theta)$, iteration $\tau$ takes the form:
\begin{equation}\label{eq:GD}
    \theta_{\tau+1}=\theta_{\tau}-\eta \nabla \mathcal{L}(\theta_\tau).
\end{equation}
}}

\textbf{Attack and defense assumptions.}
We consider a worst-case scenario, where the attacker has knowledge of the defender’s training procedure (e.g.\! learning rate, optimization algorithm), architecture, {and defense strategy}, but cannot influence training, initialization, or mini-batch sampling.
In transfer learning where the defender uses a pre-trained model and only trains the last layer, 
we assume the parameters of the pre-trained model are known to the attacker. 
However,
the defender is not aware of the target example 
or the specific patch
chosen by the attacker.
We also assume that the defender does not have access to additional clean data points.

\begin{figure*}[t]
\vskip 0.2in
\begin{center}
    \begin{subfigure}[b]{0.29\textwidth}
    \includegraphics[width=1\textwidth]{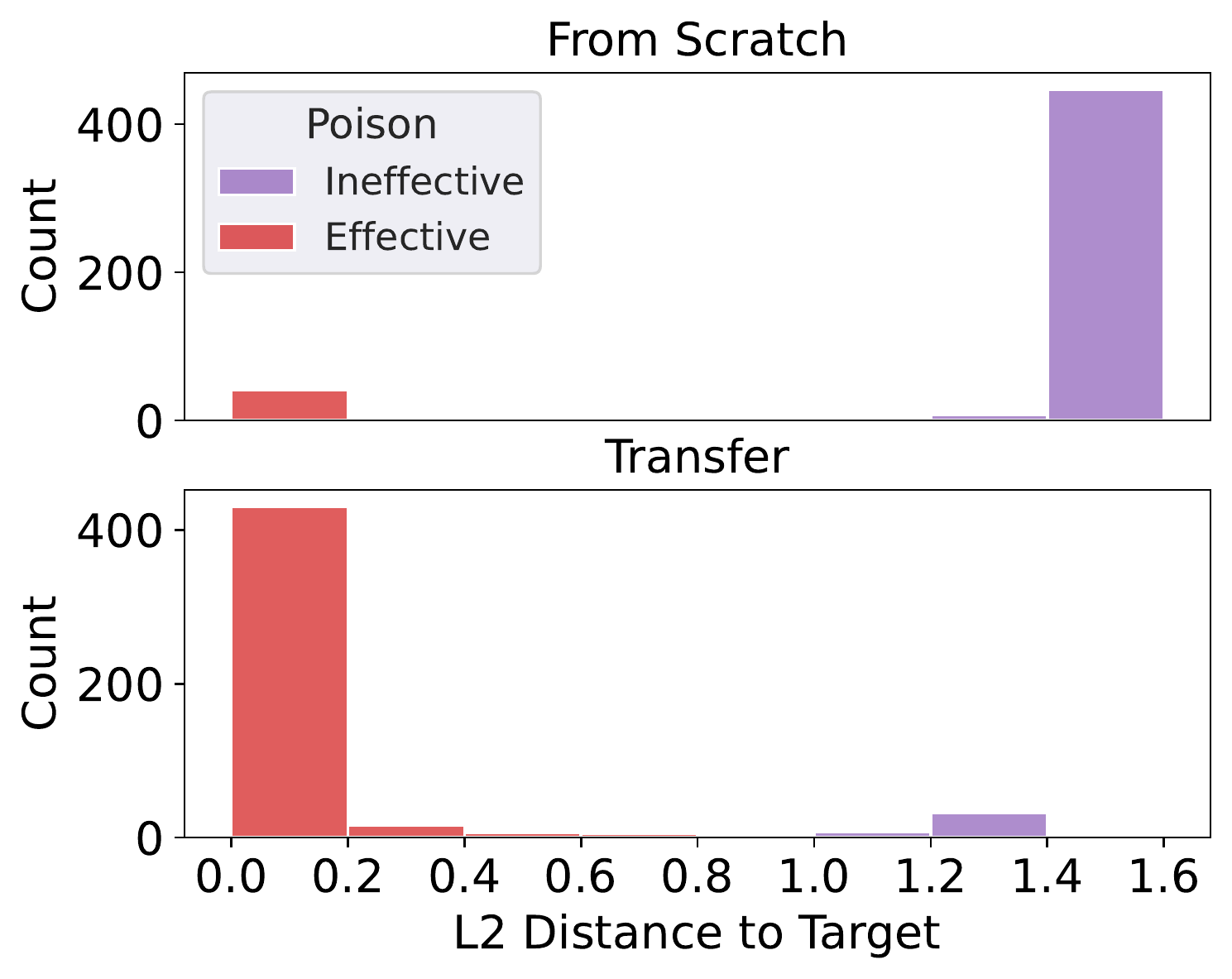}\caption{Poison distribution}\label{fig:eff_hist}
    \end{subfigure}\hfill
    \begin{subfigure}[b]{.70\textwidth}
    \includegraphics[width=\textwidth]{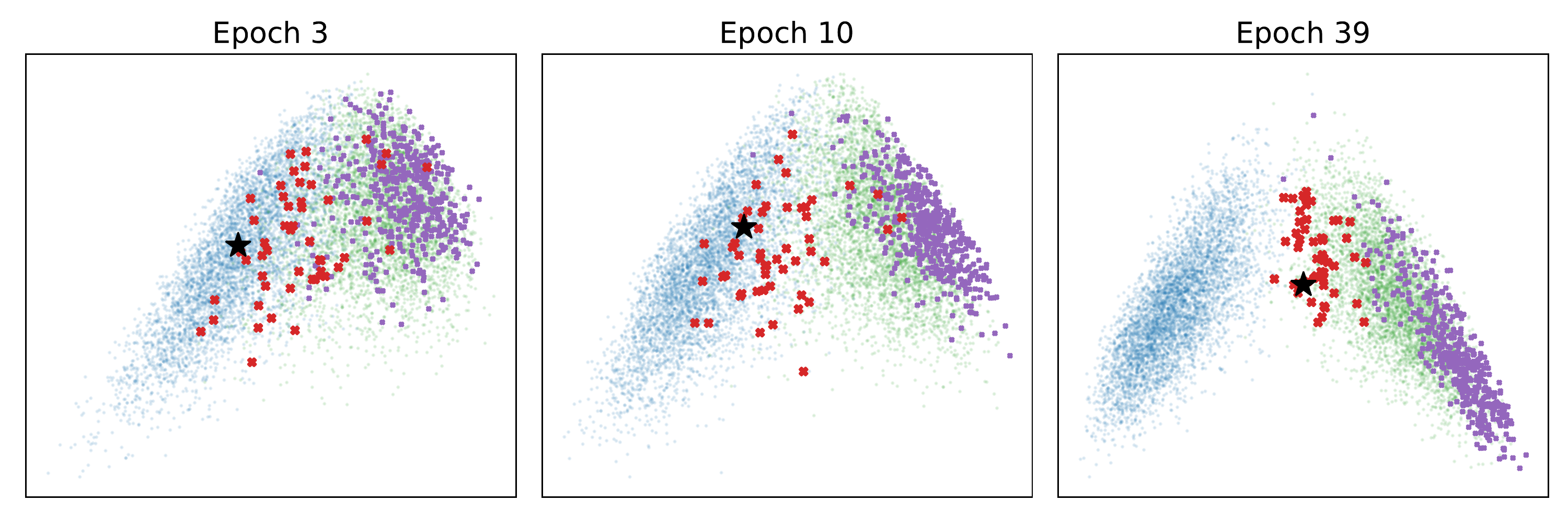}
    \caption{Training from scratch for 40 epochs with GM poisons}\label{fig:pca}
    \end{subfigure}
\vspace{-3mm}
\caption{500 effective (red) and ineffective (purple) poisons crafted by GM and BP in from-scratch and transfer learning scenarios on CIFAR10. (a) Number of effective vs. ineffective poisons and their distance to the target in the (last layer) gradient space of a clean model; (b) Embeddings of effective (red) and ineffective (purple) poisons, and clean examples of the target (blue) and poison (green) class, projected on the first 2 principal components. Effective poisons are not examples with the lowest confidence or highest loss. }
\label{fig:dist_conf}
\end{center}
\vskip -0.2in
\end{figure*}

\vspace{-2mm}
\subsection{Motivation}
For a {targeted} poisoning attack to be successful, the target needs to be misclassified as the adversarial class $y_\text{adv}$. 
Effectively, the poisons need to pull the representation of the target toward the poison class.
To do so, they need to mimic the gradient of the adversarially labeled target. Formally,
\begin{equation}\label{eq:gm}
    \nabla\mathcal{L}(x_t,y_{\text{adv}},\theta)\approx\frac{1}{|V_p|}\sum_{i\in V_p} \nabla\mathcal{L}(x_i+\delta_i,y_i,\theta)
\end{equation} 
needs to hold for any $\theta$ encountered during training.

This is the motivation behind the poison generation in the end to end training scenario. 
In particular, 
Gradient Matching \citep{geiping2021witches} and Sleeper Agent \cite{souri2021sleeper} explicitly minimize the alignment (cosine similarity) between poison and target gradient as in Eq. \eqref{eq:gm}, 
using a clean pre-trained model. 
Although the poisons are generated using a pre-trained clean model, \citep{geiping2021doesn} empirically showed that the alignment between the gradient of adversarial and training loss remains large during the training.
%
MetaPoison \citep{huang2020metapoison} uses a number of partially-trained models to generate poisons that minimize the adversarial loss at different stages
during the training. 
Bullseye Polytope \citep{aghakhani2021bullseye} maximizes the similarity between representations of the poisons and target. In doing so, it implicitly minimizes the alignment between poison and target gradients w.r.t. the penultimate layer, which captures most of the gradient norm variation \cite{katharopoulos2018not}.

In the transfer learning scenario, the poisons are crafted to have a similar representation to that of the target. Here, a linear layer is trained on the poisoned data using the representations obtained from a pre-trained  clean model. The gradient of the linear model is proportional to the representations learned by the pre-trained model.
Therefore, by maximizing the similarity between the representations of the poisons and the adversarially labeled target, the attack indeed increases the alignment between their gradients. 

Crucially, the better the poisons can surround the target in the gradient space, the more effective the attack becomes. This is demonstrated by the superior success rate of Bullseye Polytope \cite{aghakhani2021bullseye} and Convex Polytope \cite{zhu2019transferable}, compared to that of Feature Collision  \cite{Shafahi2018poisonfrogs}. While Feature Collision only optimizes the poisons to have a similar representation to that of the adversarially labeled target, Convex Polytope moves poisons until the target is inside their convex hull, and Bullseye Polytope makes further refinements to move the target away from the polytope boundary.

\subsection{Not all the poisons are created equal}
To successfully prevent poisoning attacks, we make the following key observation:
Not all the poisoned examples are responsible for the success of the attack. 
We define \textit{effective poisons} as examples that make the attack successful. That is, if the model is trained with effective poisons, the attack will be successful even if all the other poisons are removed. 
In contrast, if the effective poisons are eliminated, the remaining (ineffective) poisons cannot make the attack successful. Fig. \ref{fig:eff_hist} shows 500 effective and ineffective poisons generated by Gradient Matching (GM) and Bulleyes Polytope (BP) in the training from scratch and transfer learning scenarios. 
We tried different combinations of the poisons and identified the smallest subset of poisons that is responsible for the success of the attack. We observe that indeed not all poisons are effective.
While 
for from-scratch training 
only 8\% of the poisons are effective, for 
transfer learning around 90\% of the poisons are effective. 

We explain the above observation as follows: not all the randomly selected examples can be modified by bounded perturbations to have a gradient that closely matches that of the target. 
When training from scratch, attacks can only craft a handful of effective poisons as the poisons need to match the very high-dimensional gradient of the target with bounded perturbations. 
On the other hand, during transfer learning, poisons are optimized to match the much lower-dimensional gradient of the target. Hence, attacks can craft a much larger number of effective poisons. 

\begin{figure}[t]
\vskip 0.2in
\begin{center}
\vspace{-5mm}
\centerline{
\includegraphics[width=\columnwidth]{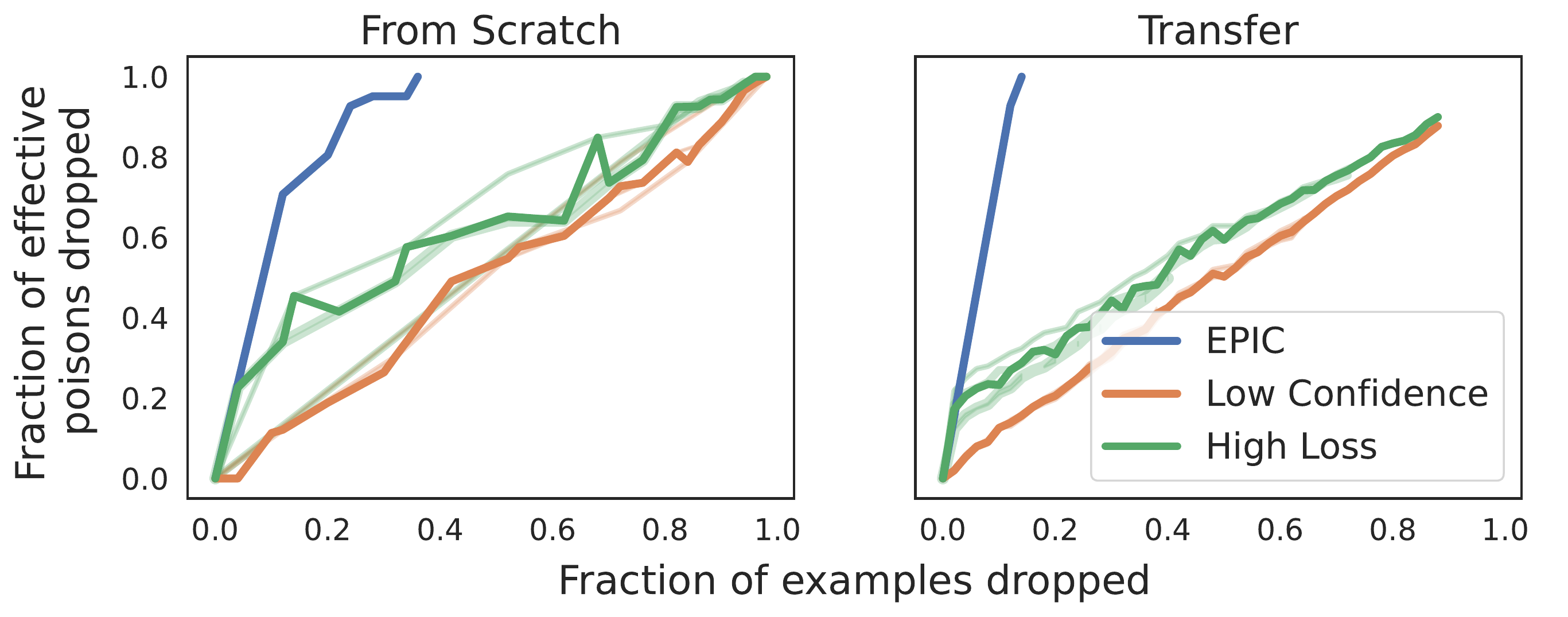}
}\vspace{-3mm}
\caption{Fraction of effective poisons dropped vs fraction of all examples dropped during training on CIFAR10 poisoned with GM, for our method (\alg) vs lowest-confidence and highest-loss with thresholds .25,.5/1,2 shown by transparent colors, and their average shown in opaque. Left: from scratch. Right: transfer learning. 
}
\vspace{-10mm}
\label{fig:num_dropped}
\end{center}
\end{figure}

\subsection{Effective poisons are not examples with highest loss or lowest confidence}\vspace{-1mm}
It is important to note that effective poisons are not the data points around the decision boundary for which the model is not confident, or 
outliers that have a higher loss than other data points in their class. 
Fig. \ref{fig:pca} shows the embedding of clean and poisoned examples of the poison and target class during training from scratch. We see that effective poisons can be within the poison or target class, or at the boundary of the classes, at different training iterations.
Fig. \ref{fig:num_dropped} shows the fraction of effective poisons eliminated when we drop examples with the highest loss or lowest confidence with various thresholds during the training. 
We see that dropping lowest-confidence or highest-loss examples {during the training} indeed removes a larger number of \textit{clean} data points, and cannot successfully eliminate the effective poisons. 



\begin{figure*}
    \centering
    \begin{subfigure}{0.285\textwidth}
        \includegraphics[width=\textwidth]{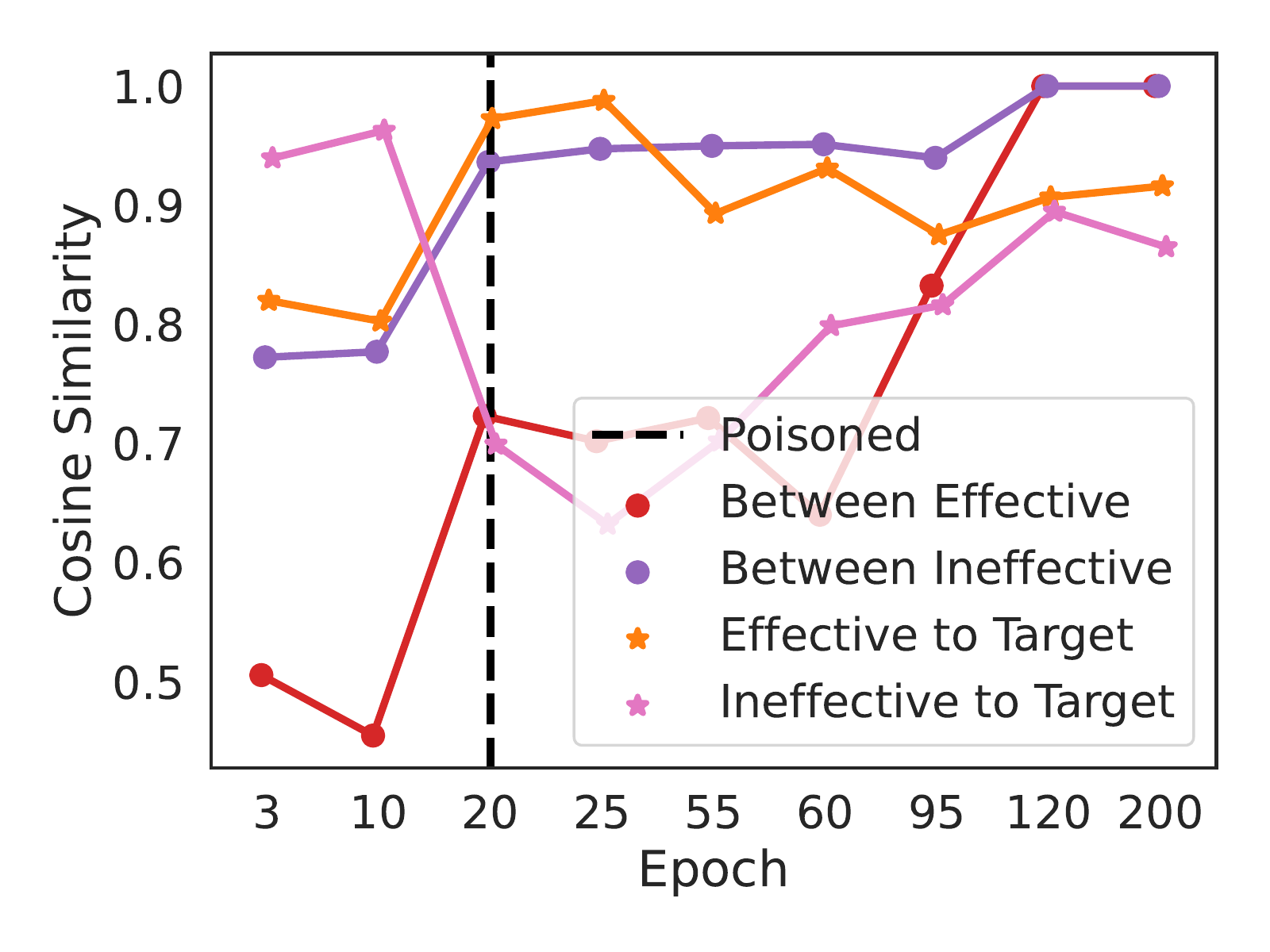}
        \caption{Cosine similarity between effective/ineffective poisons and the target. }\label{subfig:cosine}
    \end{subfigure}\hfill
    \begin{subfigure}{0.32\textwidth}
        \includegraphics[width=\textwidth]{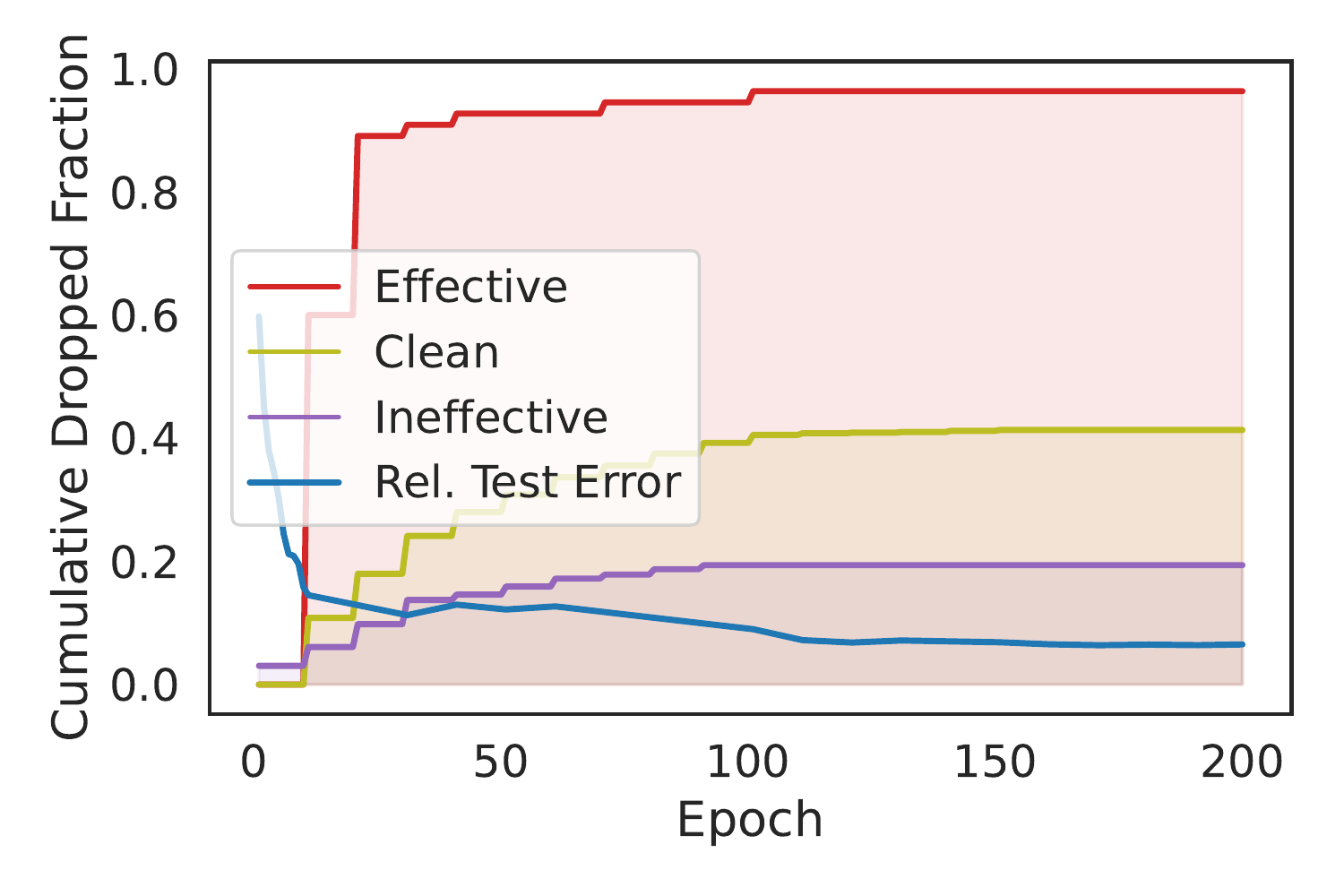}
        \caption{\!Cumulative fraction of (in)effective poisons \& cleans dropped by \alg vs test error. 
    }\label{subfig:cumulative}
    \end{subfigure}\hfill
    \begin{subfigure}{0.32\textwidth}
    \includegraphics[width=\textwidth]{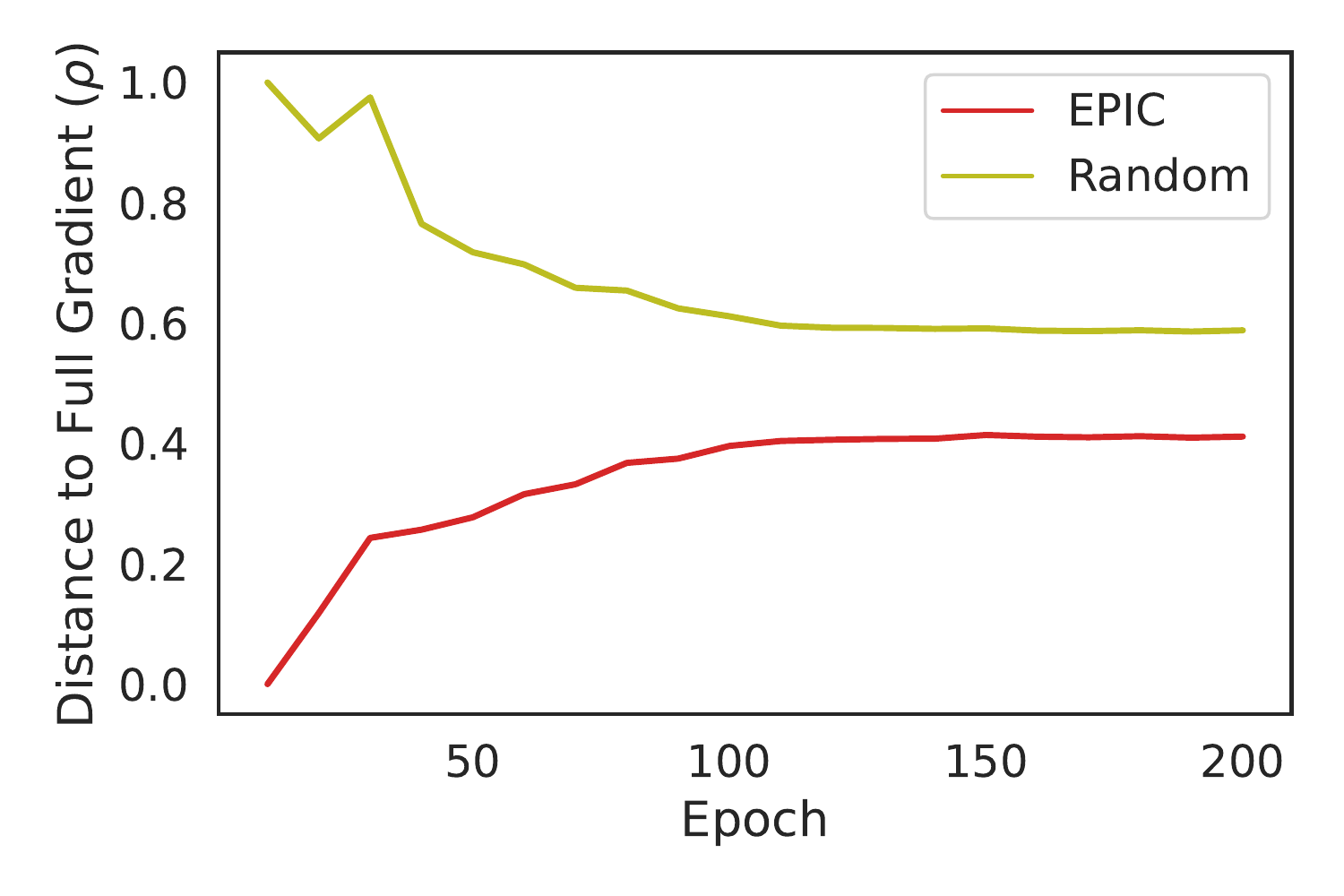}
    \caption{Gradient difference of examples not dropped with \alg~\!vs random, \!with full data.}\label{subfig:gradient}
    \end{subfigure}
    \vspace{2mm}
    \caption{Training with \alg~on CIFAR10 poisoned with GM. (a) The similarity between effective poisons' gradients to each other becomes small (they get isolated) after the warmup period, (b) \alg~effectively eliminates effective poisons while dropping a smaller fraction of clean examples, (c) \alg~preserves main gradient components, hence remaining examples have a closer gradient to that of the full data, compared to random subsets of the same size. Thus, \alg~preserves the training dynamics. 
    }
    \label{fig:compare_full}
\end{figure*}

\subsection{Effective poisons become isolated in gradient space}

Attacks exploit the non-convex nature of the neural network loss to optimize poisons that match the target gradient.
For ineffective poisons, attacks cannot successfully modify the base example with bounded perturbations to match the target gradient. This is the case where loss is relatively smooth in a ball of radius $\epsilon$ around the base.
Thus, for ineffective poisons, attacks can only increase the alignment between the gradients of ineffective poisons with that of the target to some extent. 
In doing so, 
the similarity between ineffective poisons' gradients becomes larger. 
Hence, they form larger gradient clusters in the poison class, as shown by Fig. \ref{fig:pca}.

On the other hand, effective poisons can be modified under bounded perturbations to match the target gradient. This is the case where there are sharp regions in a ball of radius $\epsilon$ around a base example. 
Here, 
the base can be perturbed and taken to such sharp regions, and its gradient can be further optimized there to match the target gradient.
During the training on the poisoned data, the gradients of effective poisons move far away from their class and get close to the target. 
But, they each have a \textit{different trajectory} (starting from different base examples) for interpolating between their base and the target gradients. 
These trajectories are neither similar to each other (as they start from different bases) nor similar to other examples in the base class (as they end up matching the target's gradient in another class).
Fig. \ref{subfig:cosine} shows that while the similarity between gradients of effective poisons and target increases during the training, the gradient of effective poisons is very different from each other
after a few epochs of training, {and before the model gets poisoned}.
Hence, effective poisons' gradients become isolated in the gradient space, early in training. \looseness=-1
%
Such isolated points in low-density gradient regions can be best identified by proximity-based methods, such as $k$-medoids, 
as we discuss next. 




\subsection{Eliminating the effective poisons}
To prevent data poisoning while maintaining the generalization performance of the network, we aim at identifying and removing the effective poisons. 
To do so, our key idea is to drop data points that have a different gradient compared to other examples in their class, i.e., are isolated in the gradient space during training. 
As we discuss next, dropping such points effectively eliminates the majority of poisoning attacks 
with only a slight impact on the gradient of the full training loss. 
By 
preserving the important gradient components,
we guarantee similar training dynamics and convergence to a close neighborhood of the solution obtained by training on full data.
\begin{figure*}[h]
    \begin{subfigure}[b]{0.3\textwidth}
         \centering
         \includegraphics[width=\textwidth]{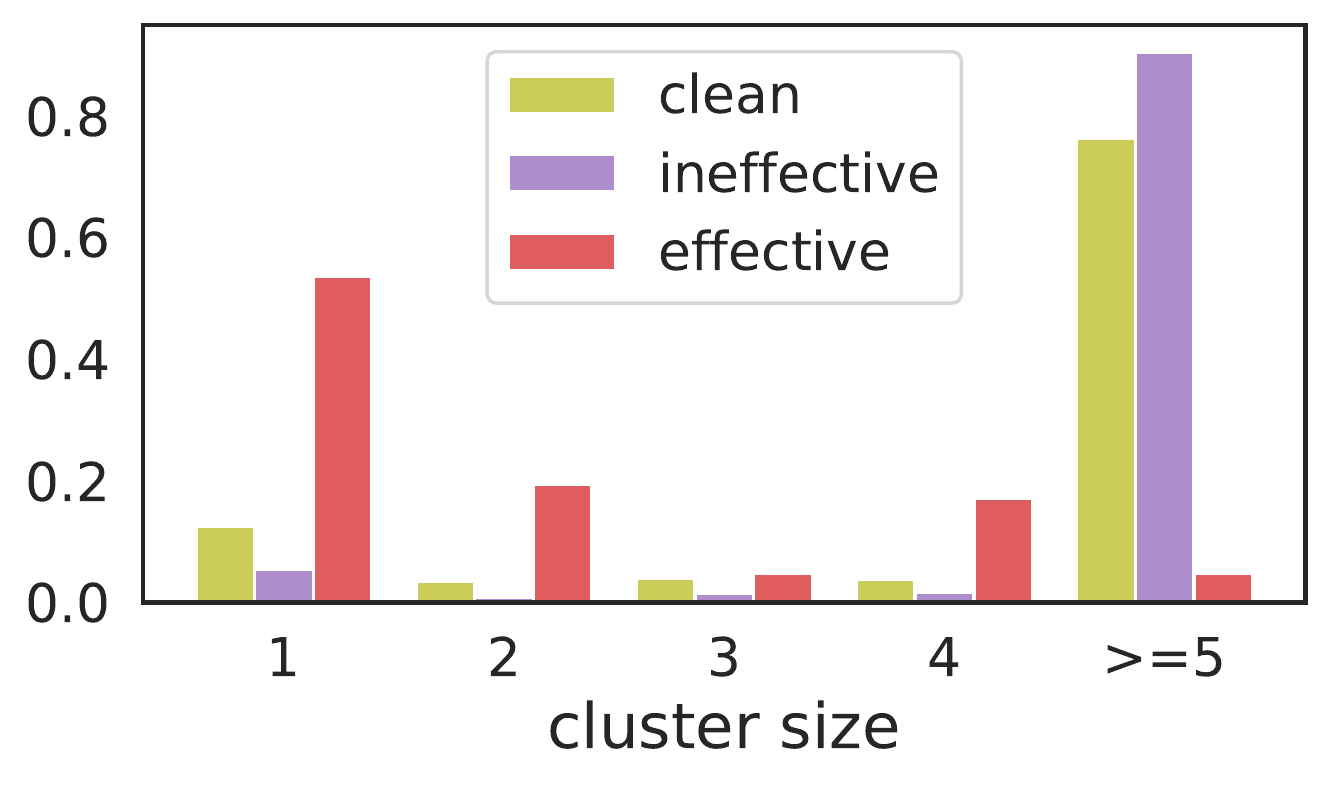}
         \caption{{Epoch 10. Most of the effective poisons are isolated as clusters of size 1.}}
         \label{fig:gm-10}
     \end{subfigure}
     \hfill
     \begin{subfigure}[b]{0.3\textwidth}
         \centering
         \includegraphics[width=\textwidth]{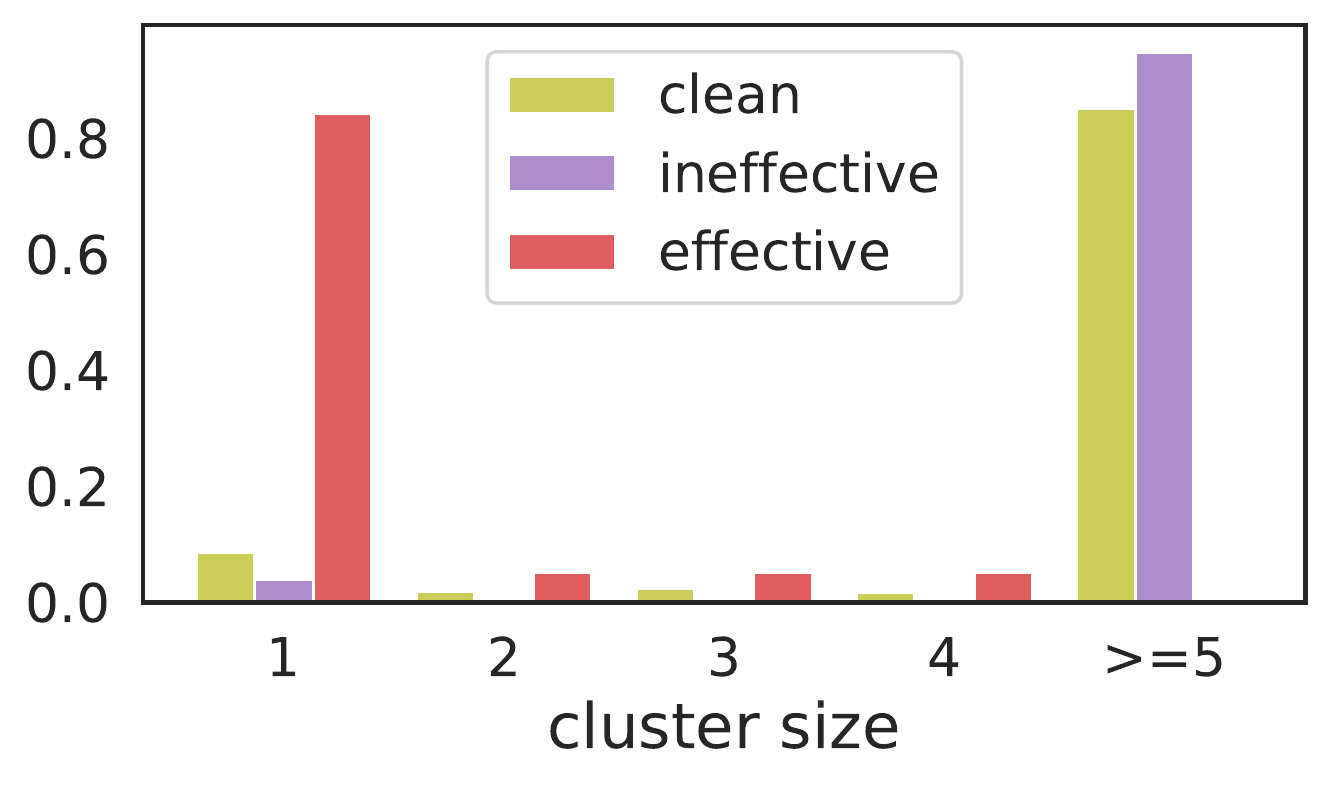}
         \caption{{Epoch 20. Most of remaining effective poisons become isolated.}}
         \label{fig:gm-20}
     \end{subfigure}
     \hfill
     \begin{subfigure}[b]{0.3\textwidth}
         \centering
         \includegraphics[width=\textwidth]{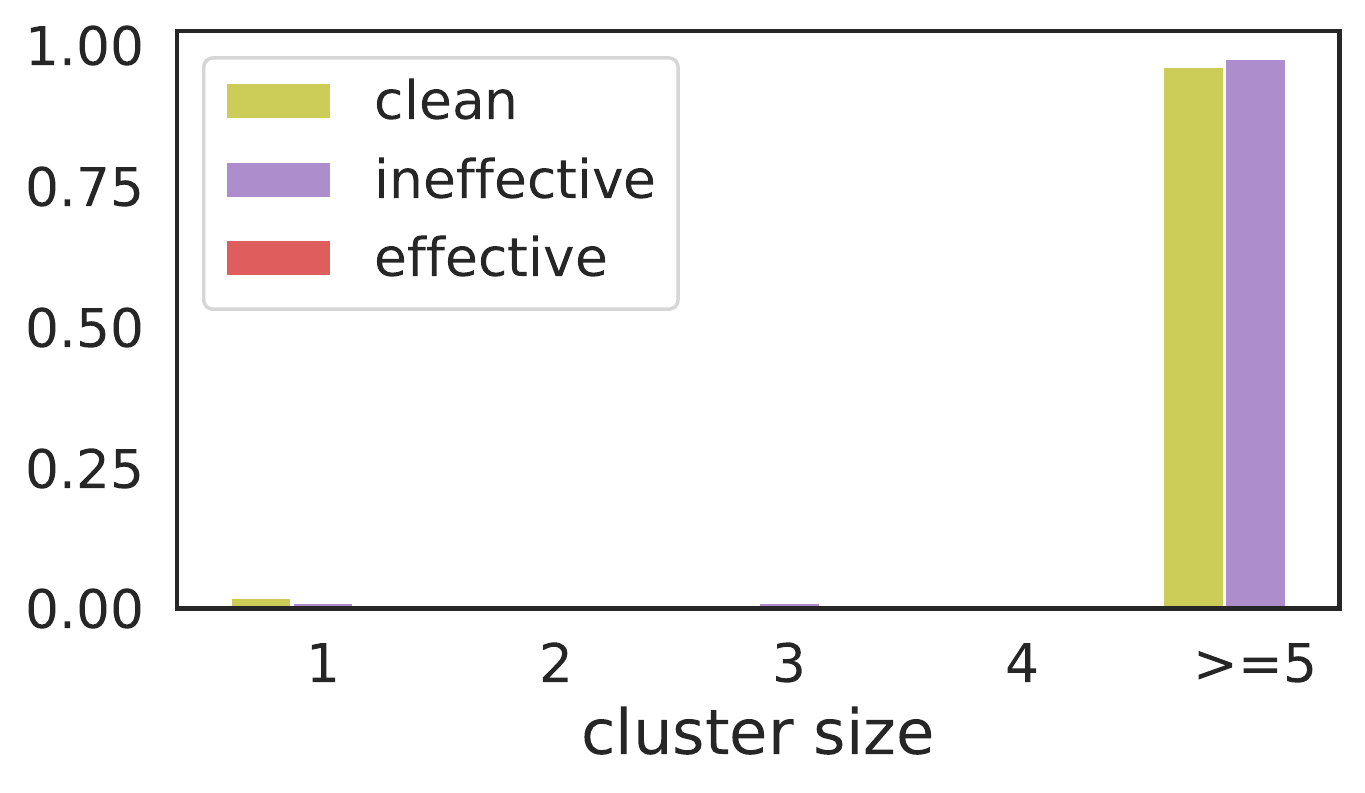}
         \caption{{Epoch 80. Clean examples form larger gradient clusters during the training.}}
         \label{fig:gm-60}
     \end{subfigure}
     \vskip -0.15in
        \caption{{Fraction of clean vs Gradient Matching poisons in gradient clusters of different sizes, during from-scratch learning with {\alg} for 200 epochs. Effective poisons become isolated during training and can be iteratively eliminated by {\alg}.} }
        \label{fig:gm-defend}
    \vskip 0.15in
    \begin{subfigure}[b]{0.3\textwidth}
         \centering
         \includegraphics[width=\textwidth]{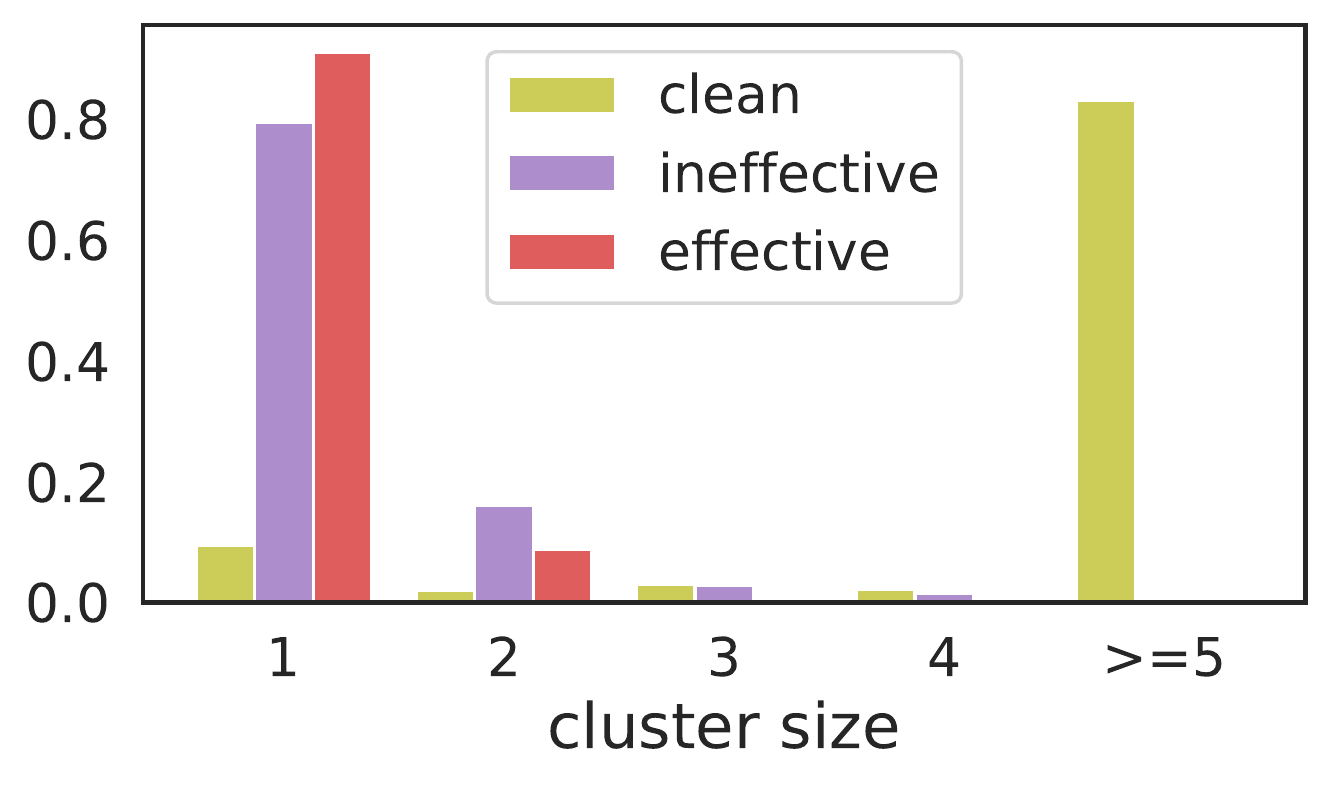}\vspace{-1mm}
         \caption{Epoch 1. {Most of the effective poisons are isolated as clusters of size 1.}}
         \label{fig:bp-1}
     \end{subfigure}
     \hfill
     \begin{subfigure}[b]{0.3\textwidth}
         \centering
         \includegraphics[width=\textwidth]{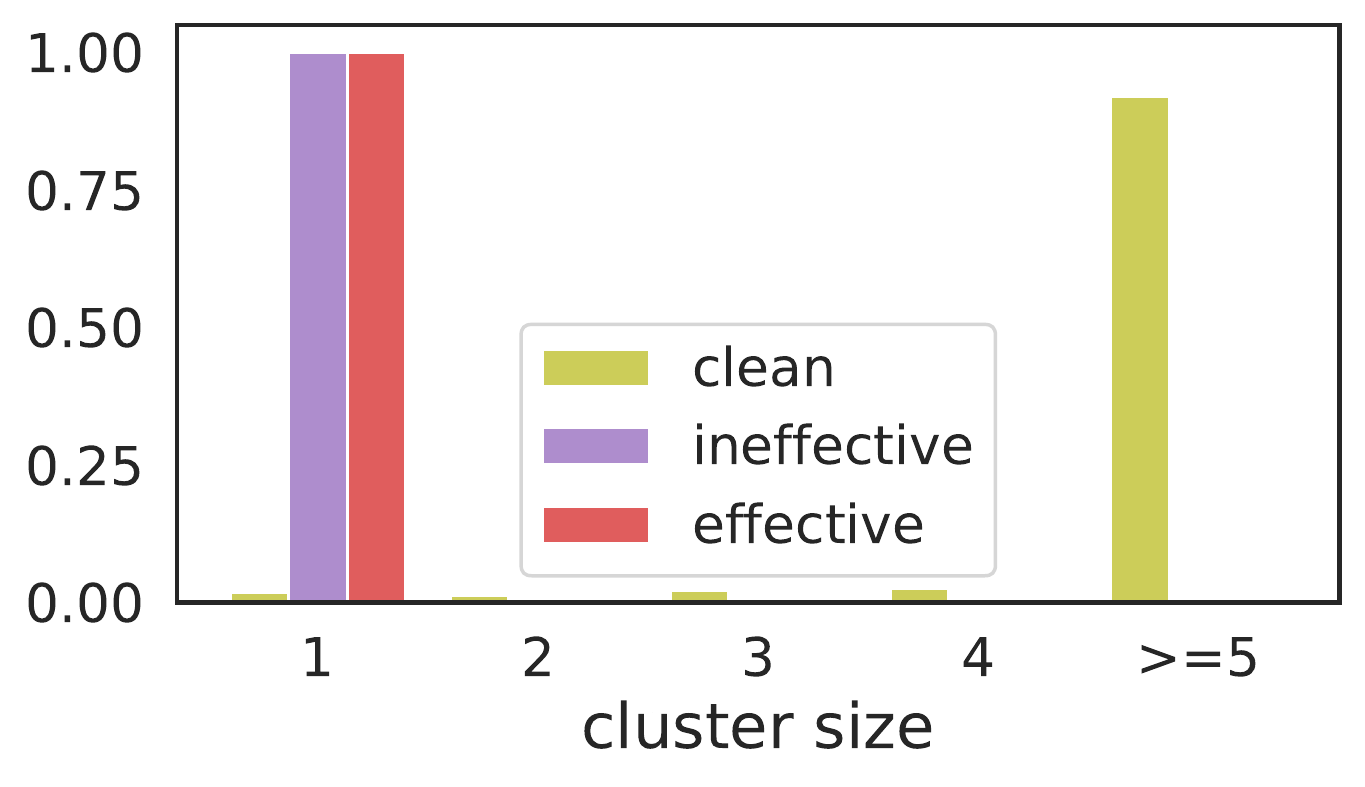}\vspace{-1mm}
         \caption{Epoch 2. Remaining effective poisons become isolated as clusters of size 1.}
         \label{fig:bp-2}
     \end{subfigure}
     \hfill
     \begin{subfigure}[b]{0.3\textwidth}
         \centering
         \includegraphics[width=\textwidth]{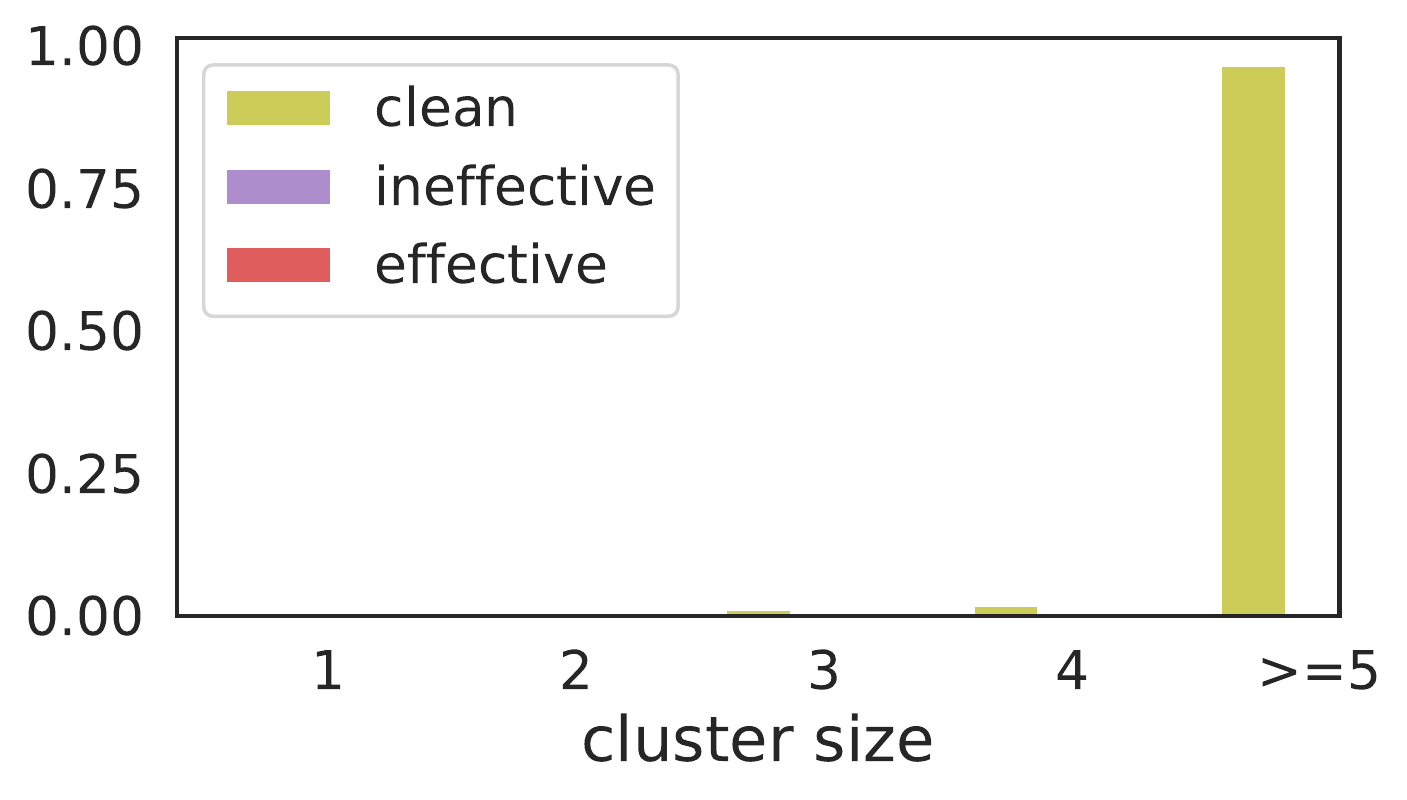}\vspace{-1mm}
         \caption{Epoch 20. {Clean examples form larger gradient clusters during the training.}}
         \label{fig:bp-30}
     \end{subfigure}
     \vskip -0.15in
        \caption{{Fraction of clean vs Bullseye Polytope poisons in gradient clusters of different sizes}, during transfer learning with \alg~for 40 epochs. Effective poisons become isolated during training and can be iteratively eliminated by \alg. }
        \label{fig:bp-defend}
        \vspace{-3mm}
\end{figure*}
%
\begin{algorithm}[tb]
   \caption{
   Effective Poison Identification
   (\alg)}
   \label{alg}
\begin{algorithmic}
   \STATE {\bfseries Input:} Training data indexed by $V$, submodular facility location function $F$, loss function $\mathcal{L(\cdot)}$, warmup iterations $K$, poison drop interval $T$, {number of medoids $k$}.
   \STATE {\bfseries Output:} Subset $S \subseteq V$
   \STATE {Train the network for $K$ epochs on $V$.}
   \FOR{every $T$ epochs}
   \FOR{examples $V_c$ in class $c\in [C]$}
   \STATE Initialize $S \leftarrow \emptyset, Z \leftarrow \emptyset $
   \WHILE{$|S| < k/C$}
   \STATE $j \in \arg\max_{e\in V_c\backslash S}F(e|S)$
   \STATE $S = S \cup \{j\}$
   \ENDWHILE
   \FOR{$j=1$ {\bfseries to} $|S|$}
   \STATE $\!\!\!\gamma_j \!\!=\!\!\! \sum_{i \in V_c}\!\!\mathbb{I}[j\!\!=\!\!\arg\!\min_{s\in S}\!||\nabla^L_f \mathcal{L}_i(\theta) \!-\!\! \nabla^L_f  \mathcal{L}_{s}(\theta)||]$ 
   \IF{$\gamma_i == 1$}
   \STATE $Z = Z \cup \{j\}$ 
   \ENDIF
   \ENDFOR
   \STATE ${V} = V \setminus Z$
   \ENDFOR
   \STATE {Train on $V$ for $T$ epochs.}
   \ENDFOR
\end{algorithmic}
\end{algorithm}
To find the effective poisons that do not have a similar gradient to the other data points in their class, we train the model for a few epochs (warmup). Then, we
iteratively find and drop the isolated points in low-density gradient regions.
To do so, we find the \textit{medoids}---the most centrally located data points---of each class, in the gradient space. Then, we assign every data point to its closest medoid, and drop the medoids to which no other data point is assigned. 
In our experiments, we show that selecting small subsets (10\%-30\% of data) of medoids at every iteration
can successfully prevent various types of data poisoning attacks. 
Fig. \ref{fig:gm-defend}, \ref{fig:bp-defend} show that during training from scratch or transfer learning,
effective poisons are isolated medoids of the gradient space. 
Hence, our strategy successfully identifies the majority of the effective poisons in both from scratch and transfer learning settings, while only dropping  a small number of clean examples (Fig.\! \ref{fig:num_dropped},\! \ref{subfig:cumulative}). \looseness=-1

The set of medoids of a class minimizes the average gradient dissimilarity to all the other data points in the class. For a specific value of $k$, the set of $k$-medoids can be found as: 
\begin{align}\label{eq:medoid}
    S^*_\tau \in & {{\arg\min}_{\substack{S\subseteq V\\|S|\leq k}}}\sum_{i \in V} \min_{j \in S}
    \|\nabla \mathcal{L}_i(\theta_\tau)-\nabla \mathcal{L}_j(\theta_\tau)\|_2,  
\end{align}
where $\mathcal{L}_i(\theta)=\mathcal{L}(x_i,y_i,\theta)$ is the loss associated with (potentially poisoned) training example $i\in V$.
The minimization problem (\ref{eq:medoid}) is NP-hard. However, it can be turned into 
maximizing a submodular\footnote{
A set function $F:2^V \rightarrow \mathbb{R}^+$ is submodular if $F(S\cup\{e\}) - F(S) \geq F(T\cup\{e\}) - F(T),$ for any $S\subseteq T \subseteq V$ and $e\in V\setminus T$.  
$F$ is \textit{monotone} if $F(e|S)\geq 0$ for any $e \! \in \! V \! \setminus \! S$ and $S \subseteq V$.} 
facility location objective:
\begin{align}\label{eq:fl_min}
    \!\!\!S^*_\tau &\in {\arg\min}_{\substack{S\subseteq V}} |S|, \quad s.t.\\ 
    &F(S)=\sum_{i \in V}\max_{j \in S} c_0 - \|\nabla \mathcal{L}_i(\theta_\tau)-\nabla \mathcal{L}_j(\theta_\tau)\|_2,\nonumber
\end{align}
where $c_0$ is a constant satisfying $c_0 \geq \|\nabla \mathcal{L}_i(\theta_\tau)-\nabla \mathcal{L}_j(\theta_\tau)\|_2$, for all $i,j \in V$.
For maximizing a monotone submodular function, the greedy algorithm provides a $(1-1/e)$ approximation guarantee \cite{wolsey1982analysis}. The greedy algorithm starts with the empty set $S_0=\emptyset$, and at each iteration $t$, it chooses an element $e\in V$ that maximizes the marginal utility $F(e|S_{t})=F(S_{t}\cup\{e\}) - F(S_{t})$. Formally,
$S_t = S_{t-1}\cup\{{\arg\max}_{e\in V} F(e|S_{t-1})\}$.
The computational complexity of the greedy algorithm is $\mathcal{O}(nk)$. 
However, its complexity can be reduced to $\mathcal{O}(|V|)$ using stochastic methods \cite{mirzasoleiman2015lazier}, and can be further improved using lazy evaluation \cite{minoux1978accelerated} and distributed implementations \cite{mirzasoleiman2013distributed}. 

During the training, the gradients of data points change at every iteration. To identify the effective poisons,
we need to update the gradient medoids iteratively. 
The gradient vectors can be very high-dimensional, in particular when training from scratch. To efficiently solve Eq. \eqref{eq:fl_min}, we rely on a recent result showing that the variation of the gradient norms is mostly captured by the gradient of the loss w.r.t. the input to the last layer \cite{katharopoulos2018not}. 
Hence, upper-bound on the normalized difference between pairwise gradient dissimilarities can be efficiently calculated:
\begin{align}
    \| \nabla \mathcal{L}_i(\theta_\tau) -& \nabla \mathcal{L}_j(\theta_\tau) \|_2  
     \leq     \mathcal{O}\big(\|\nabla^{L}_f\mathcal{L}_{i}(\theta_\tau) \!-\! \nabla^{L}_f\mathcal{L}_{j}(\theta_\tau) \|_2\big)\nonumber
\end{align}
where $\nabla^{L}_f\mathcal{L}_{i}$ is gradient of the loss function $\mathcal{L}$  w.r.t. the input to the last layer $L$ for data point $i$. 
%
The above upper bound is marginally more expensive to calculate than loss. 
Hence, upper bounds on the gradient dissimilarities can be efficiently calculated. Alg. \ref{alg} illustrates the pseudocode.

Iteratively eliminating the isolated medoids during the training allows us to successfully prevent various types of attacks. 
At the same time, as \alg~drops scattered gradient outliers and doesn't skew larger (main) gradient clusters, 
it only introduces a small limited error ($\rho$) on the full training gradient. 
Fig. \ref{subfig:gradient} shows that the gradient of the remaining training examples during training with \alg~is much closer to the full training gradient, compared to that of random subsets of the same size. 
Theorem \ref{thm:thm} leverages this idea to
upper-bound the difference between the loss of the model trained with \alg~and the model trained on the full data,
at every step of training. This ensures similar training (loss) dynamics to that of training on the full data,
and allows the network to obtain a similar generalization performance. 


\begin{table*}[h]
\caption{Average attack success rate and validation accuracy for \alg~against various data poisoning attacks (200-epoch pipeline). 
}
\vspace{-3mm}
\label{tab:attacks}
\vskip 0.15in
\begin{center}
\begin{small}
\begin{sc}
\begin{tabular}{lccccr}
\toprule
Attack & Senario & \multicolumn{2}{c}{Undefended} & \multicolumn{2}{c}{Defended} \\ 
 & & Att Succ.$\uparrow$ & Test Acc.$\uparrow$ & Att Succ.$\downarrow$ & Test Acc.$\uparrow$\\
\midrule
Gradient Matching & from-scratch & 45\% & 94.95\% & 1\%  & 90.26\%\\
Sleeper Agent (backdoor) & from-scratch & 78.54\% & 94.42\% & 11.55\%  & 88.28\%\\
\midrule
Bullseye Polytope & transfer & 86\% & 94.69\% & 1\% & 94.80\%\\ 
Feature Collision & transfer & 40\% & 94.68\% & 0\% & 94.81\%\\
\midrule
Bullseye Polytope & finetune & 80\% & 92.24\% & 0\% & 92.38\%\\ 
\bottomrule
\end{tabular}
\end{sc}
\end{small}
\end{center}
\vskip -0.2in
\end{table*}

\begin{theorem}
\label{thm:thm}
Assume that the loss function $\mathcal{L}(\theta)$ is $\mu$-PL$^*$ on a set $\Theta$, i.e., $\frac{1}{2}\| \nabla \mathcal{L}(\theta)\|^2\geq \mu \mathcal{L}(\theta), \forall \theta\in \Theta$.
Assume $\rho$ is the maximum change in the gradient norm due to dropping points. Then, applying gradient descent with a constant learning rate $\eta$ 
has similar training dynamics to that of training on the full data. I.e.,
\begin{align}
\label{eq:convergence}        
\mathcal{L}(\theta_{t}) \leq (1-{\eta\mu} {})^{t} \mathcal{L}(\theta_0) - \frac{1}{2\mu}(\rho^2-2\rho \nabla_{\max}).        
\end{align}
\end{theorem}
\vspace{-3mm}
The proof can be found in the Appendix.

Compared to existing defense strategies, our method does not require a pre-trained clean model, is not attack-specific, can be applied very efficiently during the training, and provides a quality guarantee for the performance of the trained model.

\vspace{-1mm}
\subsection{Adaptive attacks}
Adaptive attacks can generate more powerful poisons by taking into account the knowledge of the particular defense mechanisms in place. For example, Gradient Matching \cite{geiping2021witches} and Sleeper Agent \cite{souri2021sleeper} include augmented data points that are transformed with e.g. crop and flip in addition to the original ones during poison crafting in Eq. \eqref{eq:gm}. In doing so,  the attack can successfully poison the model even when data augmentations like crops and flips are applied to the learning pipeline.
For adaptive attacks to be successful in presence of \alg, they need to generate clusters of effective poisons.
To do so, the attacker may craft poisons with similar  gradient trajectories during the training, or optimize the choice of base examples that result in clustered poison gradients. However, 
crafting poisons with similar gradient trajectories during the training makes the poison optimization prohibitive and may result in less effective attacks.
While selecting similar base images does not lead to clustered effective poisons
due to non-convex nature of loss, 
optimizing the choice of base examples worth further investigation in future work. 

Next, we show that our method achieves superior performance compared to existing defense techniques.

%% file: experiments.tex
\vspace{-2mm}
\section{Experiments}

\subsection{Against Data Poisoning Attacks}
We evaluate the effectiveness of defense methods against data poisoning attacks, during \textit{from-scratch training}, \textit{transfer learning} and \textit{fine-tuning}. For our evaluation, we use the standardized data poisoning benchmark~\cite{schwarzschild2020just}, with 200 training epochs, starting learning rate of 0.1, and a decaying factor of 10 at epochs 100, 150. 
As several defense methods are prohibitive to be applied to a standard learning pipeline with 200 epochs, we also consider a \textit{proxy} setup used by \cite{geiping2021doesn} which trains for only 40 epochs, with a starting learning rate of 0.1 and decaying factor of 10 at epochs 25, 35.
\vspace{-2mm}
\begin{table*}[h]
\caption{Avg. Poison Success versus validation accuracy for various defenses against the gradient matching attack of \cite{geiping2021witches} in the from-scratch setting. The proposed Robust Training Against Data Poisoning is listed as \alg. 
}
\vspace{-3mm}
\label{tab:defenses}
\vskip 0.15in
\begin{center}
\begin{small}
\begin{sc}
\begin{tabular}{lccccr}
\toprule
Epoch & Defense & Attack Succ.$\downarrow$ & Test Acc.$\uparrow$ & Time(hr:min)\\
\midrule
40 & None & 25\% & 92.48\% & 00:15 \\
\midrule
40 & DeepKNN~\cite{peri2020deep} & 21\% & 91.86\%  & 02:25 \\
40 & Spectral Signatures~\cite{tran2018spectral} & 17\% & 90.13\% & 00:40\\
40 & Activation Clustering~\cite{chen2019detecting} & 9\% & 84.20\% & 00:31\\
40 & Diff. Priv. SGD~\cite{hong2020effectiveness} & 2\% & 70.34\% & 00:16\\
\midrule
40 & Adv. Poisoning-0.25~\cite{geiping2021doesn} & 4\% & 91.48\% & 01:53  \\
40 & Adv. Poisoning-0.5~\cite{geiping2021doesn} & 1\% & 90.67\% & 02:02  \\
40 & Adv. Poisoning-0.75~\cite{geiping2021doesn} & 0\% & 87.97\% & 02:26  \\
\midrule
40 & \alg-0.1 (Proposed) & 2.7\%$\pm$0.6\% & 90.92\%$\pm$0.26\%& 00:22 \\
40 & \alg-0.2 (Proposed) & 1.3\%$\pm$0.6\% & 88.95\%$\pm$0.08\% & 00:19 \\
40 & \alg-0.3 (Proposed) & 1.0\%$\pm$0.0\% & 87.03\%$\pm$0.11\% & 00:17 \\
\midrule
\midrule
200 & None & 45\% & 94.95\% & 01:18  \\
\midrule
200 & Spectral Signatures~\cite{tran2018spectral} & 10\% & 92.99\% &  03:22\\
200 & Activation Clustering~\cite{chen2019detecting} & 11\% & 90.88\% & 02:33 \\
200 & Diff. Priv. SGD~\cite{hong2020effectiveness} & 2\% & 80.71\% & 01:23 \\
\midrule
200 & \alg-0.1 (Proposed) & 2.3\%$\pm$0.6\% & 92.50\%$\pm$0.03\%& 01:50\\
200 & \alg-0.2 (Proposed) & 1.0\%$\pm$1.0\% & 89.71\%$\pm$0.06\%& 01:35\\
200 & \alg-0.3 (Proposed) & 0.7\%$\pm$0.6\% & 87.05\%$\pm$0.05\%& 01:28\\
\bottomrule
\end{tabular}
\end{sc}
\end{small}
\end{center}
\vskip -0.2in
\end{table*}

 \begin{table}[h]
 \vspace{-4mm}
\caption{Defending against the BP attack on TinyImageNet while training from scratch. Our method (\alg) can train more accurate models than the SOTA defense (AP) without increasing the success rate of poisoning attacks, and is more scalable. 
}\vspace{-3mm}
\label{tab:imagenet}
\vskip 0.15in
\begin{center}
\begin{small}
\begin{sc}
\begin{tabular}{lccl}
\toprule
Defense & Attack Succ.$\downarrow$ & Test Acc.$\uparrow$ & Time$\downarrow$\\
\midrule
\midrule
None & 40\% & 61.80\% & 1hr\\
AP{-0.5} & 0\% & 53.54\% & 7hrs\\
\alg-0.2 & 0\% & 57.50\% & 1hr\\
\bottomrule
\end{tabular}
\end{sc}
\end{small}
\end{center}
\vspace{-7mm}
\end{table}


\subsubsection{From-Scratch Training}
 \vspace{-2mm}
We model the from-scratch training experiments based on the benchmark setting~\cite{schwarzschild2020just}. For our attack model, we select 1\% of the training examples as poisons, which are perturbed within the $l_\infty$ ball of radius $\epsilon=8/255$. The defender initializes a model based on a random seed and trains on the poisoned dataset using SGD. To maximize reproducibility, we only use publicly available poisoned datasets generated by authors of the attacks.

Unless otherwise specified, we augment training images with random horizontal flip followed by random cropping, 
and per-channel normalization. For our proposed defense, we
{run {\alg} with $T\!=\!2$ in a 40-epoch training pipeline, or $T\!=\!10$ in a 200-epoch training pipeline. }
 
{\textbf{Warmup} The more medoids we select for each class, the longer warmup period (K) we need for \alg. In the experiments, we set $K=10$ for \alg-0.1, $K=20$ for \alg-0.2 and $K=30$ for \alg-0.3.} 
 \vspace{-3mm}
\paragraph{Gradient Matching (GM)} GM is currently the state-of-the-art among data poisoning attacks for from-scratch training \cite{schwarzschild2020just}. \cite{geiping2021witches} shows it significantly outperform the other effective attack, MetaPoison~\cite{huang2020metapoison}.
We test 100 datasets provided by the authors
. 
The datasets were generated based on the 100 preset benchmark settings, each with 500 specific bases and a target image. 
 %
 We follow the training hyperparameters specified by the benchmark to train  ResNet-18 from scratch with 128 examples per mini-batch.
Table \ref{tab:attacks} shows that the average attack success rate of GM on these 100 datasets is 45\% and the average test accuracy is 94.95\%. We see that our proposed defense, \alg, is able to successfully drop the average attack success rate to only 1\% while keeping the test accuracy above 90\%. 

\vspace{-3mm}
\paragraph{Bullseye Polytope (BP)}{ \citet{schwarzschild2020just} shows the superiority of BP attack in training VGG models~\cite{simonyan2014very} from scratch on TinyImageNet, a subset of the ILSVRC2012 classification dataset~\cite{deng2009imagenet}. We tested the first 10 example datasets provided by the benchmark, 
and observed an attack success rate of 40\%. 
Training with {\alg} drops the attack success rate significantly to 0\%, as shown in Table \ref{tab:imagenet}. 
}

\vspace{-4mm}
\paragraph{{Sleeper Agent (SA)}} SA is the only backdoor attack that can achieve a higher than single-digit success rate on CIFAR-10 in the from-scratch setting. We generated 20 poisoned datasets with SA ($\epsilon=16$) using the source-target class pairs in the first 20 CIFAR-10 benchmark settings. For backdoor attacks, we evaluate the attack success rate over 1000 patched test images. The average attack success rate is 78.54\% without defenses and 11.55\% with \alg.
 


 \subsubsection{Transfer Learning}
 \vspace{-2mm}
 Here, we use the 40-epoch pipeline of \cite{geiping2021doesn} to evaluate defense methods.
 The same pretrained model is used for generating the attack and for transfer learning onto the defender. Similar to the from-scratch setting, the attacker can modify 1\% of 50000 training examples in CIFAR10 with $\epsilon = 8$. The linear layer (classifier) of the pretrained model is then re-initialized and trained with the poisoned dataset with all the other layers (feature extractor) fixed during the training.
 This white-box setup allows attacks to produce stronger poisons.
Here, we apply $\alg$ with $T\!=\!1$.\looseness=-1

\vspace{-3mm}
 \paragraph{Bullseye Polytope (BP)} According to~\cite{schwarzschild2020just}, Bullseye Polytope attack~\cite{aghakhani2021bullseye} has the highest average attack success rate in the white-box setting. When evaluated on all 100 benchmark setups, BP succeeded in 86 of them. Table \ref{tab:attacks} shows that $\alg$ could successfully drop the attack success rate to only 1\% while even increasing the test accuracy of the model.
 
 \vspace{-3mm}
 \paragraph{Feature Collision (FC)}
 As imposing the $l_{\infty}$ constraint $\epsilon=8$ will greatly reduce the power of Feature Collison attack~\cite{schwarzschild2020just}, we keep the $l_2$ regularization term in their original optimization objective to 
impose a soft rather than hard constraint on the $l_{\infty}$ perturbation.
 We generate 20 poisoned datasets using the first 20 benchmark setups (indexed from 0 to 19). 
 With the default seed used by the benchmark, the attack success rate of these 20 datasets generated by FC is 40\% before and 0\% after we apply our $\alg$ defense, as shown in Table \ref{tab:attacks}.
 

\vspace{-2mm}
\subsubsection{Finetuning}
 \vspace{-2mm}
{We also consider the finetuning scenario in which the classifier is re-initialized and the feature extractor is not fixed during the training. We follow the same setup in \cite{geiping2021doesn}, test 20 datasets poisoned with BP, and report the result in Table \ref{tab:attacks}. Again, {\alg} successfully prevents all attacks in this scenario without decreasing the test accuracy.}

\vspace{-3mm}

\subsection{Comparison to SOTA Defenses against GM}
\vspace{-2mm}
Table~\ref{tab:defenses} compares the effectiveness of our model with existing defense methods against the state-of-the-art GM attack, in both 200-epoch and 40-epoch training scenarios.
We see that $\alg$ can successfully drop the success rate of GM while allowing the model to achieve superior performance. 
We note that, unlike existing defense methods, our method is easily scalable to standard deep learning pipelines. 

\vspace{-4mm}
\paragraph{Scalability}
As many defense methods~\cite{geiping2021doesn,peri2020deep} are prohibitive when applied to the standard 200-epoch pipeline under time or space constraints, they are evaluated using a 40-epoch pipeline. However, 
as Table~\ref{tab:defenses} shows,  
training a model on the same poisoned datasets 
for more epochs increases attack success rate. Therefore, defenses that are successful within 40 epochs are not guaranteed to have the same 
effectiveness when models are trained for longer.
On the contrary, our proposed defense requires nearly no extra time compared to normal training. Time spent on running \alg~ every few epochs is usually well compensated by training time saved every epoch on the examples we drop. 
Table~\ref{tab:defenses} includes the time for each defense on CIFAR10 poisoned with GM, and Table ~\ref{tab:imagenet} compares \alg~with AP on TinyImagenet poisoned with BP.
We report the wall-clock time of training a model with each defense on a single NVIDIA A40 GPU with 4 workers. 
We see that \alg~effectively reduces various attacks' success rates while having substantially faster run time.

\begin{figure}[t]
\centering
    \centering
         \includegraphics[width=0.7\columnwidth]{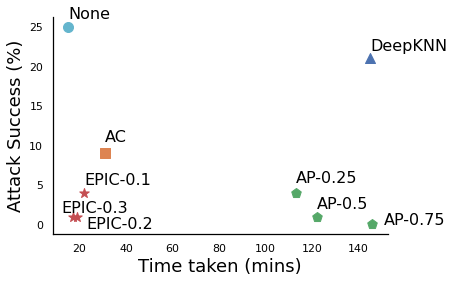}
        \vspace{-4mm}
         \caption{Attack success rate vs. running time of different defenses, for {GM attack on CIFAR-10 with} the 40 Epochs pipeline.}
         \label{fig:tradeoff}
         \vspace{-6mm}
\end{figure}

\vspace{-4mm}
\paragraph{Strength of Defense}
Due to computational constraints and the scalability problem mentioned above, we only scale the two general adversarial training methods, Adversarial Training~\cite{madry2018towards} and DP-SGD~\cite{hong2020effectiveness} to the standard 200-epoch training pipeline. According to Table~\ref{tab:defenses}, \ref{tab:imagenet} and Fig.~\ref{fig:tradeoff}, our method provides the best trade-offs between the defended attack success rate and the overall test accuracy. Adversarial Poisoning~\cite{geiping2021doesn} can give equally good trade-offs but requires 6x training time. Other defenses either cannot guarantee a low attack success rate or have a high computation cost.

 \begin{table}[t]
\caption{Comparison 
of avg. poison accuracy, validation accuracy and time against the strongest attack GM \cite{geiping2021witches} with $\epsilon=16$ in the from-scratch setting for 40 epochs. Our proposed defense is listed as \alg. 
}\vspace{-3mm}
\label{tab:ap}
\vskip 0.15in
\begin{center}
\begin{small}
\begin{sc}
\begin{tabular}{lcr}
\toprule
Defense & Attack Succ.$\downarrow$ & Test Acc.$\uparrow$\\

\midrule
\midrule
None & 90\% & 92.01\%\\
\midrule
AP{-0.25} & 35\% & 91.21\% \\
AP{-0.5} & 10\% & 90.58\% \\
AP{-0.75} & 0\% & 87.97\% \\
\midrule
\alg-0.1 & 10\% & 91.15\%\\
\alg-0.2 & 0\% & 89.07\%\\
\bottomrule
\end{tabular}
\end{sc}
\end{small}
\end{center}
\vspace{-5mm}
\end{table}

\subsection{Comparison under Larger Perturbations}
\vspace{-2mm}
Attacks usually have higher success rates when allowed to perturb the base images within a larger $\epsilon$ constraint~\cite{schwarzschild2020just}. 
We generate 20 poisoned datasets with a larger $\epsilon=16$
 with GM. We use the default 20 seeds used in~\cite{geiping2021witches} to sample 500 base and 1 target images from CIFAR10. Table \ref{tab:ap} shows that for larger $\epsilon$, \alg~ achieves a superior performance compared to the strongest baseline, adversarial poisoning.


%% file: conclusion.tex
\section{Conclusion}
\vspace{-2mm}
We proposed an efficient defense mechanism against various data poisoning attacks. 
We showed that under bounded perturbations, only a small number of poisons can be optimized to have a gradient that is close enough to that of the target and make the attack successful. Such examples move away from their original class and get isolated in the gradient space.
Consequently, we showed that training on large gradient clusters of each class successfully eliminates the effective poisons, and guarantees similar training dynamics to that of training on the full data.
Our experiments showed that our method significantly decreases the success rate of the state-of-the-art targeted attacks, including Gradient Matching, Bullseye Polytope.
We note that our method is the only effective defense against strong poisoning attacks, which easily scales to standard deep learning pipelines.

%% file: appendix.tex
\section{Proof of Theorem \ref{thm:thm}}
A loss function $\mathcal{L}(w)$ is considered $\mu$ -PL on a set $\mathcal{S}$, if the following holds:
\begin{align}
    \frac{1}{2}\|\mathbf{g}\|^{2} \geq \mu\left(\mathcal{L}(w)-\mathcal{L}\left(w_{*}\right)\right), \forall w \in \mathcal{S}
    \label{eq:pl}
\end{align}
where $w_{*}$ is a global minimizer.
When additionally $\mathcal{L}\left(w_{*}\right) = 0$, 
the $\mu$-$\text{PL}$ condition is equivalent to the $\mu$-$\text{PL}^{*}$ condition 
\begin{align}
    \frac{1}{2}\|\mathbf{g}\|^{2} \geq \mu\mathcal{L}(w), \forall w \in \mathcal{S}.
\end{align}

For Lipschitz continuous $\mathbf{g}$ and $\mu$-PL$^*$ condition, gradient descent on the entire dataset yields
\begin{align}
    \mathcal{L}(w_{t+1}) - \mathcal{L}(w_t) \leq -\frac{\eta} {2}\|\mathbf{g}_t\|^2 \leq -\eta\mu \mathcal{L}(w_t),\label{eq:grad_rate}
\end{align}
and,

\begin{align}
    \mathcal{L}(w_{t})\leq(1-\eta\mu)^t \mathcal{L}(w_0),\label{eq:lip}
\end{align}
which was shown in \cite{liu2020toward}.
We build upon this result.

    For the subset we have
    \begin{align}
        \mathcal{L}(w_{t+1}) - \mathcal{L}(w_t) 
        &\leq -\frac{\eta}{2}\|\mathbf{g}_t^S\|^2
    \end{align}    
    By substituting Eq. (\ref{eq:grad_rate}) we have.
    \begin{align}
        &\leq -\frac{\eta}{2}(\|\mathbf{g}_t\|-\rho)^2\\
        &= -\frac{\eta} {2}(\|\mathbf{g}_t\|^2+\rho^2-2\rho\|\mathbf{g}_t\|)\label{eq:pre_spectral_upper}\\
        &\leq - \frac{\eta} {2}(\|\mathbf{g}_t\|^2+\rho^2-2\rho \nabla_{\max})\\
        &\leq -\frac{\eta} {2}(2\mu \mathcal{L}(w_t)+\rho^2-2\rho \nabla_{\max}) \label{eq:pl_applied}
    \end{align}
    where we can upper bound the norm of $\mathbf{g}_t$ in Eq. (\ref{eq:pre_spectral_upper}) by a constant $\nabla_{\max}$. And Eq. (\ref{eq:pl_applied}) follows from the $\mu$-PL condition from Eq. (\ref{eq:pl}). While the loss is very non-convex during the first part of training, it becomes nearly convex afterward \cite{fort2020deep}.  \alg~starts dropping points after a few epochs of training, where Lipschitzness, $\mu$-PL condition, and norm-bounded gradients are likely to hold. In Eq. \eqref{eq:lip}, LHS directly results from Lipschitzness \cite{boyd2004convex}.

    Hence,
    \begin{align}
        \mathcal{L}(w_{t+1}) \leq (1-{\eta\mu} ) \mathcal{L}(w_t) - \frac{\eta} {2}(\rho^2-2\rho \nabla_{\max})
    \end{align}
    Since, $\sum_{j=0}^k(1-{\eta\mu} {})^j\leq\frac{1}{\eta\mu}$, for a constant learning rate $\eta$ we get
    \begin{align}
        \mathcal{L}(w_{t+1}) \leq (1-{\eta\mu} {})^{t+1} \mathcal{L}(w_0) - \frac{1}{2\mu}(\rho^2-2\rho \nabla_{\max})
    \end{align}


%% file: icml_poisoning.bbl
\begin{thebibliography}{40}
\providecommand{\natexlab}[1]{#1}
\providecommand{\url}[1]{\texttt{#1}}
\expandafter\ifx\csname urlstyle\endcsname\relax
  \providecommand{\doi}[1]{doi: #1}\else
  \providecommand{\doi}{doi: \begingroup \urlstyle{rm}\Url}\fi

\bibitem[Abadi et~al.(2016)Abadi, Chu, Goodfellow, McMahan, Mironov, Talwar,
  and Zhang]{abadi2016deep}
Abadi, M., Chu, A., Goodfellow, I., McMahan, H.~B., Mironov, I., Talwar, K.,
  and Zhang, L.
\newblock Deep learning with differential privacy.
\newblock In \emph{Proceedings of the 2016 ACM SIGSAC conference on computer
  and communications security}, pp.\  308--318, 2016.

\bibitem[Aghakhani et~al.(2021)Aghakhani, Meng, Wang, Kruegel, and
  Vigna]{aghakhani2021bullseye}
Aghakhani, H., Meng, D., Wang, Y.-X., Kruegel, C., and Vigna, G.
\newblock Bullseye polytope: A scalable clean-label poisoning attack with
  improved transferability.
\newblock In \emph{2021 IEEE European Symposium on Security and Privacy
  (EuroS\&P)}, pp.\  159--178. IEEE, 2021.

\bibitem[Borgnia et~al.(2021)Borgnia, Cherepanova, Fowl, Ghiasi, Geiping,
  Goldblum, Goldstein, and Gupta]{borgnia2021strong}
Borgnia, E., Cherepanova, V., Fowl, L., Ghiasi, A., Geiping, J., Goldblum, M.,
  Goldstein, T., and Gupta, A.
\newblock Strong data augmentation sanitizes poisoning and backdoor attacks
  without an accuracy tradeoff.
\newblock In \emph{ICASSP 2021-2021 IEEE International Conference on Acoustics,
  Speech and Signal Processing (ICASSP)}, pp.\  3855--3859. IEEE, 2021.

\bibitem[Boyd et~al.(2004)Boyd, Boyd, and Vandenberghe]{boyd2004convex}
Boyd, S., Boyd, S.~P., and Vandenberghe, L.
\newblock \emph{Convex optimization}.
\newblock Cambridge university press, 2004.

\bibitem[Chen et~al.(2019)Chen, Carvalho, Baracaldo, Ludwig, Edwards, Lee,
  Molloy, and Srivastava]{chen2019detecting}
Chen, B., Carvalho, W., Baracaldo, N., Ludwig, H., Edwards, B., Lee, T.,
  Molloy, I., and Srivastava, B.
\newblock Detecting backdoor attacks on deep neural networks by activation
  clustering.
\newblock In \emph{SafeAI@ AAAI}, 2019.

\bibitem[Chen et~al.(2017)Chen, Liu, Li, Lu, and Song]{chen2017targeted}
Chen, X., Liu, C., Li, B., Lu, K., and Song, D.
\newblock Targeted backdoor attacks on deep learning systems using data
  poisoning.
\newblock \emph{arXiv preprint arXiv:1712.05526}, 2017.

\bibitem[Cretu et~al.(2008)Cretu, Stavrou, Locasto, Stolfo, and
  Keromytis]{cretu2008casting}
Cretu, G.~F., Stavrou, A., Locasto, M.~E., Stolfo, S.~J., and Keromytis, A.~D.
\newblock Casting out demons: Sanitizing training data for anomaly sensors.
\newblock In \emph{2008 IEEE Symposium on Security and Privacy (sp 2008)}, pp.\
   81--95. IEEE, 2008.

\bibitem[Deng et~al.(2009)Deng, Dong, Socher, Li, Li, and
  Fei-Fei]{deng2009imagenet}
Deng, J., Dong, W., Socher, R., Li, L.-J., Li, K., and Fei-Fei, L.
\newblock Imagenet: A large-scale hierarchical image database.
\newblock In \emph{2009 IEEE conference on computer vision and pattern
  recognition}, pp.\  248--255. Ieee, 2009.

\bibitem[Fort et~al.(2020)Fort, Dziugaite, Paul, Kharaghani, Roy, and
  Ganguli]{fort2020deep}
Fort, S., Dziugaite, G.~K., Paul, M., Kharaghani, S., Roy, D.~M., and Ganguli,
  S.
\newblock Deep learning versus kernel learning: an empirical study of loss
  landscape geometry and the time evolution of the neural tangent kernel.
\newblock \emph{Advances in Neural Information Processing Systems},
  33:\penalty0 5850--5861, 2020.

\bibitem[Geiping et~al.(2021{\natexlab{a}})Geiping, Fowl, Somepalli, Goldblum,
  Moeller, and Goldstein]{geiping2021doesn}
Geiping, J., Fowl, L., Somepalli, G., Goldblum, M., Moeller, M., and Goldstein,
  T.
\newblock What doesn't kill you makes you robust (er): Adversarial training
  against poisons and backdoors.
\newblock \emph{arXiv preprint arXiv:2102.13624}, 2021{\natexlab{a}}.

\bibitem[Geiping et~al.(2021{\natexlab{b}})Geiping, Fowl, Huang, Czaja, Taylor,
  Moeller, and Goldstein]{geiping2021witches}
Geiping, J., Fowl, L.~H., Huang, W.~R., Czaja, W., Taylor, G., Moeller, M., and
  Goldstein, T.
\newblock Witches' brew: Industrial scale data poisoning via gradient matching.
\newblock In \emph{International Conference on Learning Representations},
  2021{\natexlab{b}}.
\newblock URL \url{https://openreview.net/forum?id=01olnfLIbD}.

\bibitem[Gu et~al.(2017)Gu, Dolan-Gavitt, and Garg]{gu2017badnets}
Gu, T., Dolan-Gavitt, B., and Garg, S.
\newblock Badnets: Identifying vulnerabilities in the machine learning model
  supply chain.
\newblock \emph{arXiv preprint arXiv:1708.06733}, 2017.

\bibitem[Hong et~al.(2020)Hong, Chandrasekaran, Kaya, Dumitra{\c{s}}, and
  Papernot]{hong2020effectiveness}
Hong, S., Chandrasekaran, V., Kaya, Y., Dumitra{\c{s}}, T., and Papernot, N.
\newblock On the effectiveness of mitigating data poisoning attacks with
  gradient shaping.
\newblock \emph{arXiv preprint arXiv:2002.11497}, 2020.

\bibitem[Huang et~al.(2020)Huang, Geiping, Fowl, Taylor, and
  Goldstein]{huang2020metapoison}
Huang, W.~R., Geiping, J., Fowl, L., Taylor, G., and Goldstein, T.
\newblock Metapoison: Practical general-purpose clean-label data poisoning.
\newblock \emph{Advances in Neural Information Processing Systems}, 33, 2020.

\bibitem[Jayaraman \& Evans(2019)Jayaraman and Evans]{jayaraman2019evaluating}
Jayaraman, B. and Evans, D.
\newblock Evaluating differentially private machine learning in practice.
\newblock In \emph{28th $\{$USENIX$\}$ Security Symposium ($\{$USENIX$\}$
  Security 19)}, pp.\  1895--1912, 2019.

\bibitem[Katharopoulos \& Fleuret(2018)Katharopoulos and
  Fleuret]{katharopoulos2018not}
Katharopoulos, A. and Fleuret, F.
\newblock Not all samples are created equal: Deep learning with importance
  sampling.
\newblock In \emph{International Conference on Machine Learning}, pp.\
  2525--2534, 2018.

\bibitem[Koh et~al.(2018)Koh, Steinhardt, and Liang]{koh2018stronger}
Koh, P.~W., Steinhardt, J., and Liang, P.
\newblock Stronger data poisoning attacks break data sanitization defenses.
\newblock \emph{arXiv preprint arXiv:1811.00741}, 2018.

\bibitem[Levine \& Feizi(2020)Levine and Feizi]{levine2020deep}
Levine, A. and Feizi, S.
\newblock Deep partition aggregation: Provable defenses against general
  poisoning attacks.
\newblock In \emph{International Conference on Learning Representations}, 2020.

\bibitem[Li et~al.(2021)Li, Lyu, Koren, Lyu, Li, and Ma]{li2021anti}
Li, Y., Lyu, X., Koren, N., Lyu, L., Li, B., and Ma, X.
\newblock Anti-backdoor learning: Training clean models on poisoned data.
\newblock \emph{Advances in Neural Information Processing Systems}, 34, 2021.

\bibitem[Liu et~al.(2020)Liu, Zhu, and Belkin]{liu2020toward}
Liu, C., Zhu, L., and Belkin, M.
\newblock Toward a theory of optimization for over-parameterized systems of
  non-linear equations: the lessons of deep learning.
\newblock \emph{arXiv preprint arXiv:2003.00307}, 2020.

\bibitem[Liu et~al.(2017)Liu, Ma, Aafer, Lee, Zhai, Wang, and
  Zhang]{liu2017trojaning}
Liu, Y., Ma, S., Aafer, Y., Lee, W.-C., Zhai, J., Wang, W., and Zhang, X.
\newblock Trojaning attack on neural networks.
\newblock 2017.

\bibitem[Ma et~al.(2019)Ma, Zhu, and Hsu]{ma2019data}
Ma, Y., Zhu, X.~Z., and Hsu, J.
\newblock Data poisoning against differentially-private learners: Attacks and
  defenses.
\newblock In \emph{International Joint Conference on Artificial Intelligence},
  2019.

\bibitem[Madry et~al.(2018)Madry, Makelov, Schmidt, Tsipras, and
  Vladu]{madry2018towards}
Madry, A., Makelov, A., Schmidt, L., Tsipras, D., and Vladu, A.
\newblock Towards deep learning models resistant to adversarial attacks.
\newblock In \emph{International Conference on Learning Representations}, 2018.

\bibitem[Minoux(1978)]{minoux1978accelerated}
Minoux, M.
\newblock Accelerated greedy algorithms for maximizing submodular set
  functions.
\newblock In \emph{Optimization techniques}, pp.\  234--243. Springer, 1978.

\bibitem[Mirzasoleiman et~al.(2013)Mirzasoleiman, Karbasi, Sarkar, and
  Krause]{mirzasoleiman2013distributed}
Mirzasoleiman, B., Karbasi, A., Sarkar, R., and Krause, A.
\newblock Distributed submodular maximization: Identifying representative
  elements in massive data.
\newblock In \emph{Advances in Neural Information Processing Systems}, pp.\
  2049--2057, 2013.

\bibitem[Mirzasoleiman et~al.(2015)Mirzasoleiman, Badanidiyuru, Karbasi,
  Vondr{\'a}k, and Krause]{mirzasoleiman2015lazier}
Mirzasoleiman, B., Badanidiyuru, A., Karbasi, A., Vondr{\'a}k, J., and Krause,
  A.
\newblock Lazier than lazy greedy.
\newblock In \emph{Twenty-Ninth AAAI Conference on Artificial Intelligence},
  2015.

\bibitem[Peri et~al.(2020)Peri, Gupta, Huang, Fowl, Zhu, Feizi, Goldstein, and
  Dickerson]{peri2020deep}
Peri, N., Gupta, N., Huang, W.~R., Fowl, L., Zhu, C., Feizi, S., Goldstein, T.,
  and Dickerson, J.~P.
\newblock Deep k-nn defense against clean-label data poisoning attacks.
\newblock In \emph{European Conference on Computer Vision}, pp.\  55--70.
  Springer, 2020.

\bibitem[Saha et~al.(2019)Saha, Subramanya, and Pirsiavash]{Saha2019htbd}
Saha, A., Subramanya, A., and Pirsiavash, H.
\newblock Hidden trigger backdoor attacks, 2019.

\bibitem[Schwarzschild et~al.(2020)Schwarzschild, Goldblum, Gupta, Dickerson,
  and Goldstein]{schwarzschild2020just}
Schwarzschild, A., Goldblum, M., Gupta, A., Dickerson, J.~P., and Goldstein, T.
\newblock Just how toxic is data poisoning? a unified benchmark for backdoor
  and data poisoning attacks.
\newblock \emph{arXiv preprint arXiv:2006.12557}, 2020.

\bibitem[Shafahi et~al.(2018)Shafahi, Huang, Najibi, Suciu, Studer, Dumitras,
  and Goldstein]{Shafahi2018poisonfrogs}
Shafahi, A., Huang, W.~R., Najibi, M., Suciu, O., Studer, C., Dumitras, T., and
  Goldstein, T.
\newblock Poison frogs! targeted clean-label poisoning attacks on neural
  networks, 2018.

\bibitem[Simonyan \& Zisserman(2014)Simonyan and Zisserman]{simonyan2014very}
Simonyan, K. and Zisserman, A.
\newblock Very deep convolutional networks for large-scale image recognition.
\newblock \emph{arXiv preprint arXiv:1409.1556}, 2014.

\bibitem[Souri et~al.(2021)Souri, Goldblum, Fowl, Chellappa, and
  Goldstein]{souri2021sleeper}
Souri, H., Goldblum, M., Fowl, L., Chellappa, R., and Goldstein, T.
\newblock Sleeper agent: Scalable hidden trigger backdoors for neural networks
  trained from scratch.
\newblock \emph{arXiv preprint arXiv:2106.08970}, 2021.

\bibitem[Steinhardt et~al.(2017)Steinhardt, Koh, and
  Liang]{Steinhardt17certified}
Steinhardt, J., Koh, P.~W., and Liang, P.
\newblock Certified defenses for data poisoning attacks, 2017.

\bibitem[Tao et~al.(2021)Tao, Feng, Yi, Huang, and Chen]{tao2021better}
Tao, L., Feng, L., Yi, J., Huang, S.-J., and Chen, S.
\newblock Better safe than sorry: Preventing delusive adversaries with
  adversarial training.
\newblock \emph{Advances in Neural Information Processing Systems}, 34, 2021.

\bibitem[Tran et~al.(2018)Tran, Li, and Madry]{tran2018spectral}
Tran, B., Li, J., and Madry, A.
\newblock Spectral signatures in backdoor attacks.
\newblock In \emph{Advances in Neural Information Processing Systems}, pp.\
  8000--8010, 2018.

\bibitem[Turner et~al.(2018)Turner, Tsipras, and Madry]{turner2018clean}
Turner, A., Tsipras, D., and Madry, A.
\newblock Clean-label backdoor attacks.
\newblock 2018.

\bibitem[Veldanda \& Garg(2020)Veldanda and Garg]{veldanda2020evaluating}
Veldanda, A. and Garg, S.
\newblock On evaluating neural network backdoor defenses.
\newblock \emph{arXiv preprint arXiv:2010.12186}, 2020.

\bibitem[Weber et~al.(2020)Weber, Xu, Karla{\v{s}}, Zhang, and
  Li]{weber2020rab}
Weber, M., Xu, X., Karla{\v{s}}, B., Zhang, C., and Li, B.
\newblock Rab: Provable robustness against backdoor attacks.
\newblock \emph{arXiv preprint arXiv:2003.08904}, 2020.

\bibitem[Wolsey(1982)]{wolsey1982analysis}
Wolsey, L.~A.
\newblock An analysis of the greedy algorithm for the submodular set covering
  problem.
\newblock \emph{Combinatorica}, 2\penalty0 (4):\penalty0 385--393, 1982.

\bibitem[Zhu et~al.(2019)Zhu, Huang, Li, Taylor, Studer, and
  Goldstein]{zhu2019transferable}
Zhu, C., Huang, W.~R., Li, H., Taylor, G., Studer, C., and Goldstein, T.
\newblock Transferable clean-label poisoning attacks on deep neural nets.
\newblock In \emph{International Conference on Machine Learning}, pp.\
  7614--7623, 2019.

\end{thebibliography}
